\documentclass[10pt,twocolumn,letterpaper]{article}

\usepackage{cvpr}              
\definecolor{cvprblue}{rgb}{0.21,0.49,0.74}
\usepackage[pagebackref,breaklinks,colorlinks,allcolors=cvprblue]{hyperref}
\usepackage[accsupp]{axessibility}
\usepackage{multirow}
\usepackage{graphicx}
\usepackage[table,xcdraw]{xcolor}
\usepackage{bbding}
\usepackage{wrapfig}
\usepackage{subcaption}
\usepackage{algorithm}
\usepackage{algpseudocode}
\usepackage{amsmath}
\usepackage{amssymb}
\usepackage{booktabs}
\usepackage{tabularx}
\usepackage{array}
\usepackage{enumitem}
\usepackage{multirow}
\usepackage{makecell}
\usepackage{url}
\usepackage{balance}
\usepackage[table,xcdraw]{xcolor}
\usepackage[accsupp]{axessibility}
\renewcommand{\thefootnote}{\fnsymbol{footnote}}
\def\paperID{16536} 
\def\confName{CVPR}
\def\confYear{2026}

\title{DirectFisheye-GS: Enabling Native Fisheye Input in Gaussian Splatting with Cross-View Joint Optimization}

\author{Zhengxian Yang$^{1,4}$\footnotemark[1], \quad  Fei Xie$^{1}$\footnotemark[1], \quad Xutao Xue$^{2}$, \quad  Rui Zhang$^{1}$, \quad Taicheng Huang$^{3}$, \\
Yang Liu$^{3}$, \quad Mengqi Ji$^{2}$, \quad Tao Yu$^{1}$\footnotemark[2] \\
  $^1$BNRist, Tsinghua University \;\;
  $^2$Beihang University \;\;
  $^3$JD.com, Beijing, China \;\;
  $^4$Shanghai AI Lab\\
\texttt{\small\{zx-yang23, xief22, zhangrui22\}@mails.tsinghua.edu.cn} \;\; 
\texttt{\small xuexutao@buaa.edu.cn} \\
\texttt{\small \{huangtaicheng.1, liuyang1605\}@jd.com} \;\; 
\texttt{\small jimengqi@buaa.edu.cn}  \;\; 
\texttt{\small ytrock@mail.tsinghua.edu.cn} \\
}

\begin{document}
\twocolumn[{%
\renewcommand\twocolumn[1][]{#1}%
\maketitle
\vspace{-2em}
\begin{center}
\includegraphics[width=\linewidth]{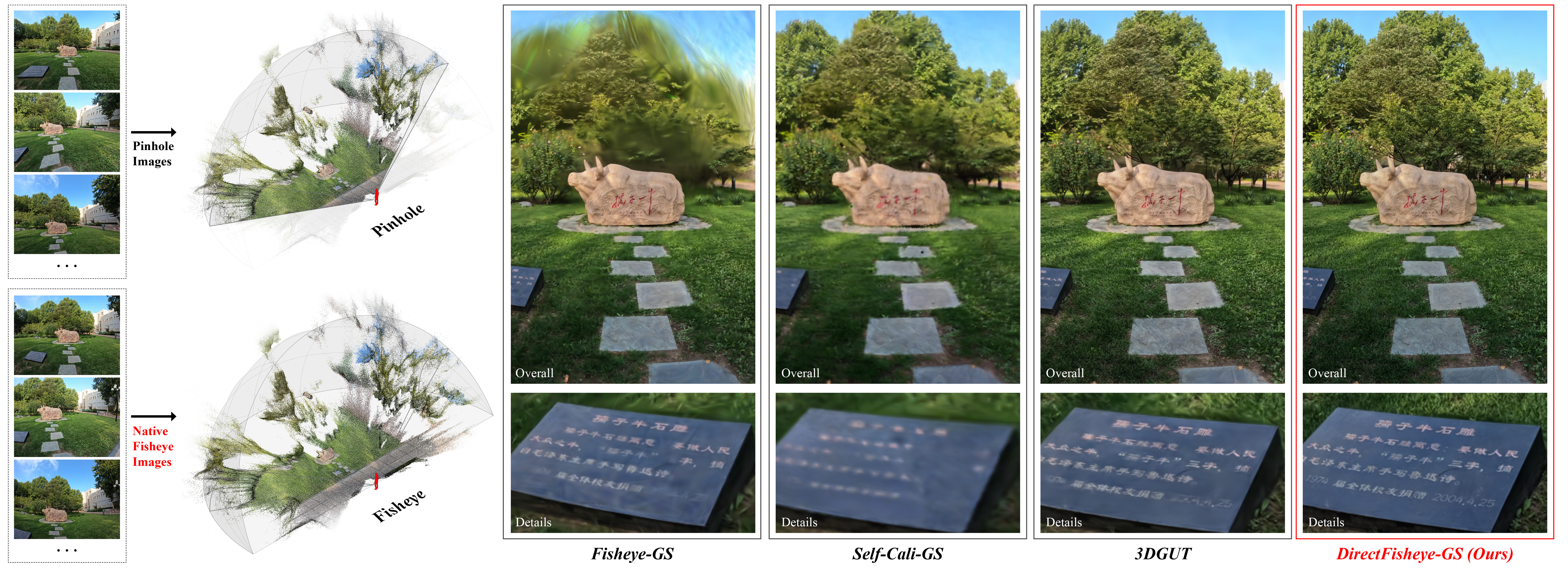}
\captionof{figure}{\textbf{DirectFisheye-GS} enables native fisheye image input for 3DGS training, avoiding information loss caused by undistortion. It achieves, and in many cases surpasses, state-of-the-art performance on various public datasets, demonstrating superior detail preservation and cross-view geometric \& global illumination consistency.}
\label{fig:teaser}
\end{center}
}]
\begin{abstract}
\vspace{-20pt}

\renewcommand{\thefootnote}{\fnsymbol{footnote}}
\footnotetext[1]{Equal contributions. \textsuperscript{$\dagger$}Corresponding author.}

3D Gaussian Splatting (3DGS) has enabled efficient 3D scene reconstruction from everyday images with real-time, high-fidelity rendering, greatly advancing VR/AR applications. Fisheye cameras, with their wider field of view (FOV), promise high-quality reconstructions from fewer inputs and have recently attracted much attention. However, since 3DGS relies on rasterization, most subsequent works involving fisheye camera inputs first undistort images before training, which introduces two problems: 1) Black borders at image edges cause information loss and negate the fisheye’s large FOV advantage; 2) Undistortion’s stretch‐and‐interpolate resampling spreads each pixel’s value over a larger area, diluting detail density— causes 3DGS overfitting these low‐frequency zones, producing blur and floating artifacts.
In this work, we integrate fisheye camera model into the original 3DGS framework, enabling native fisheye image input for training without preprocessing. 
Despite correct modeling, we observed that the reconstructed scenes still exhibit floaters at image edges: Distortion increases toward the periphery, and 3DGS's original per-iteration random-selecting-view optimization ignores the cross-view correlations of a Gaussian, leading to extreme shapes (e.g., oversized or elongated) that degrade reconstruction quality. To address this, we introduce a feature-overlap–driven cross-view joint optimization strategy that establishes consistent geometric and photometric constraints across views—a technique equally applicable to existing pinhole-camera-based pipelines. Our DirectFisheye-GS matches or surpasses state-of-the-art performance on public datasets.
\end{abstract}
\vspace{-15pt}
\section{Introduction}
\label{sec:intro}
Novel View Synthesis (NVS) is a classic and high‐profile task in computer vision, aiming to generate photorealistic images at unseen viewpoints from a set of input photos. Recent breakthroughs have drawn extensive industry interest and found wide application in immersive VR/AR, holography, tourism, and education.

The emergence of 3D Gaussian Splatting (3DGS)~\cite{kerbl20233d} marked a paradigm shift from implicit networks~\cite{mildenhall2021nerf,liu2020neural,sun2022direct, muller2022instant,yariv2023bakedsdf,sharma2024volumetric,turki2024hybridnerf} to explicit anisotropic Gaussians and rasterization-based rendering. By directly projecting 3D Gaussians onto 2D screens, 3DGS achieves real-time rendering while preserving visual fidelity. Follow-up works further optimized its scalability~\cite{kerbl2024hierarchical,liu2024cityGaussian,lin2024vastGaussian}, and memory efficiency~\cite{lee2024compact,lu2024scaffold,kheradmand20243d,mallick2024taming}. 
While 3DGS excels with pinhole cameras, its direct application to fisheye images-a cornerstone of autonomous driving, robotics, and immersive VR due to ultra-wide FOV ( $>$90°, the common ones are 120° or 180°) and efficient scene coverage with fewer captures—faces critical limitations. Fisheye lenses’ severe radial distortion and nonlinear projection violate pinhole assumptions of the original 3DGS, causing artifacts and performance degradation when naively applied.

Common workarounds involve pre-undistorting fisheye images before reconstruction, but this introduces irreversible compromises: (1) cropping distorted edges causes information loss, negating the wide-FOV advantage; (2) the stretch-and-interpolate resampling in undistortion spreads pixel values over larger areas, diluting spatial detail and leading 3DGS to overfit low-frequency regions—resulting in blurry textures and floating artifacts.
Recent studies have thus turned to directly modeling fisheye projections. Fisheye-GS~\cite{liao2024fisheye} was the first attempt to adapt 3DGS for fisheye inputs, yet it still converts images into ideal equidistant projections instead of explicitly modeling lens geometry, leading to visible distortions and floating Gaussians.
The latest and most accurate open-source solution, 3DGUT~\cite{wu20243dgut}, introduces an Unscented Transform (UT)–based formulation that augments 3DGS with ray tracing and secondary rays, achieving the current state of the art on several fisheye benchmarks. However, its sampling-based covariance propagation relies on only seven sigma points, which become insufficient under strong fisheye distortion. In high-curvature boundary regions, this sparse sampling fails to capture local anisotropy, leading to mosaic-like artifacts, slightly smoothed edges, and less precise geometry.
Moreover, both methods follow 3DGS's per-iteration random-single-view optimization strategy, which neglects correlations between overlapping viewpoints. Even with a correct fisheye camera model, insufficient optimization leads to extreme Gaussian shapes (e.g., oversized ellipsoids) and inconsistent color predictions—particularly around the boundaries of fisheye images, where distortion is most pronounced. 

To address these challenges, we propose \textbf{DirectFisheye-GS}, a novel framework that enables native fisheye inputs in 3DGS while enhancing geometric and photometric consistency through cross-view joint optimization. We integrate the Kannala-Brandt~\cite{kannala2006generic} fisheye projection model into the 3DGS pipeline, eliminating the need for undistortion preprocessing and preserving the efficiency of rasterization-based rendering. This not only ensures full compatibility with existing 3DGS viewers and commercial tools, but also retains the original spatial relationships among neighboring rays—an essential condition for multi-view consistency constraints. We then introduce a cross-view joint optimization strategy that adaptively groups training views based on feature overlap and viewpoint divergence. By simultaneously optimizing Gaussians across correlated views, our method enforces geometric consistency and mitigates shape irregularities, significantly improving reconstruction quality without sacrificing efficiency.

Our contributions are summarized as follows:
\begin{itemize}
\item \textbf{Fisheye-Compatible 3DGS:} A fisheye projection model embedded into 3DGS, enabling direct training on distorted fisheye images without preprocessing.
\item \textbf{Cross-View Joint Optimization:} A novel training strategy that leverages multi-view correlations to constrain Gaussian shapes and positions, addressing optimization ambiguity and extreme parameterization.
\item \textbf{State-of-the-Art Performance:} Extensive experiments on public datasets demonstrate that DirectFisheye-GS outperforms existing methods in both quantitative metrics (PSNR, SSIM, LPIPS) and qualitative visual fidelity, particularly in large-scale scenarios.
\end{itemize}
\section{Related Work}
\label{sec:Related Work}
\subsection{Novel View Synthesis (NVS)}
\paragraph{Neural Radiance Fields (NeRF)~\cite{mildenhall2021nerf}} introduced a groundbreaking approach to novel‐view synthesis by training an MLP that maps 3D coordinates and view directions to volume density and radiance, which are then rendered via differentiable volumetric ray marching. Subsequent work has improved rendering quality by incorporating explicit geometry priors~\cite{sun2022direct,fridovich2022plenoxels,cao2023hexplane,li2023uhdnerf} and physically based material models~\cite{haque2023instruct}, and has accelerated inference through adaptive sampling strategies~\cite{xu2022point}, hash‐based data structures~\cite{muller2022instant}, and tensor decompositions~\cite{chen2022tensorf}. Nonetheless, the underlying ray‐marching remains a bottleneck for real‐time performance, and the implicit network representation constrains scene editability. 
\vspace{-15pt}
\paragraph{3D Gaussian Splatting (3DGS)~\cite{kerbl20233d}}  
3DGS introduces an explicit anisotropic Gaussian representation combined with CUDA‐accelerated differentiable rasterization, overcoming NeRF’s efficiency limitations. By projecting 3D Gaussians directly onto the image plane, it achieves both superior rendering speed and visual fidelity, enabling real‐time, high‐quality VR/AR applications. Subsequent works have enhanced robustness to sparse views as input~\cite{chen2024mvsplat,li2024dnGaussian,zhu2024fsgs,zhang2024cor}, extended reconstruction to urban‐scale scenes~\cite{lin2024vastGaussian,liu2024cityGaussian}, and mitigated aliasing artifacts~\cite{yu2024mip,yan2024multi}. Other efforts—such as Scaffold‐GS~\cite{lu2024scaffold}, Compact3DGS~\cite{lee2024compact}, LightGaussian~\cite{fan2024lightGaussian}, and methods by Durvasula et al.~\cite{durvasula2023distwar} and Jo et al.~\cite{jo2024identifying}—focus on optimizing data structures or pruning Gaussians to reduce memory and computation.

Despite these advances, all current 3DGS pipelines share a critical limitation: they employ a NeRF‐style single‐view optimization that ignores correlations of the same Gaussian across views, leading to overfitting and extreme Gaussian shapes. MVGS~\cite{du2024mvgs} first explored multi‐view joint optimization, but still relies on randomly sampling view subsets each iteration. In this work, we propose a novel cross-view joint optimization mechanism that simultaneously considers feature overlap and viewpoint divergence to adaptively group training views, enforcing geometric and photometric consistency for each Gaussian across views and markedly improving NVS quality.
\subsection{Camera Modeling for NVS}
To overcome the limitations of existing NVS methods for nonlinear image inputs, researchers have introduced wide-FOV imaging schemes such as fisheye lens models and panoramic camera models. The ray-casting mechanism in NeRF makes it easier to extend to other camera models. SC-NeRF~\cite{jeong2021self} achieves end-to-end camera parameter optimization without feature matching by extending the fourth-order radial distortion model and introducing learnable nonlinear residual compensation. NeuroLens~\cite{xian2023neural} learns complex nonlinear distortions through bidirectional differentiable mapping. FBINeRF~\cite{wu2024fbinerf} designs an adaptive GRU module to optimize parameters for fisheye camera radial distortion. MSI-NeRF~\cite{yan2025msi} constructs an implicit radiance field using multi-spherical images, requiring only four input images for 6-DoF view synthesis. OmniNeRF~\cite{gu2022omni} integrates omnidirectional distance fields (ODF) with radiance fields, effectively mitigating edge blur. 360FusionNeRF~\cite{kulkarni360fusionnerf} and PERF~\cite{wang2024perf} employ equidistant spherical mapping and multi-view cone segmentation strategies for processing panoramic images. Although these methods have made progress in expanding camera models, they are still constrained by many limitations of the original NeRF, making them unsuitable for real-time rendering.

Recently, 3DGS has attracted researchers' attention due to its faster rendering speed and higher quality. Fisheye-GS~\cite{liao2024fisheye} is the first 3DGS work to explore nonlinear camera models. It proposes using equidistant fisheye projection mapping, but it still requires a de-distortion preprocessing step, reducing its practicality. 3DGUT~\cite{wu20243dgut} innovatively introduces Unscented Transform to model nonlinear projections, avoiding the model dependency of traditional Jacobian matrix derivation. Although it achieves a balance between accuracy and generalization, it breaks the original fully explicit architecture of 3DGS, making the training results incompatible with existing viewers and commercial pipelines. Other approaches like EVER~\cite{mai2024ever} and 3D Gaussian Ray Tracing~\cite{moenne20243d} follow NeRF-like methods to incorporate hybrid ray tracing, overcoming rasterization constraints but still facing computational bottlenecks. Self-Cali-GS~\cite{deng2025self} uses reversible residual networks to predict sparse grid displacement fields, though it suffers from slow convergence and loss of high-frequency details. 360-GS~\cite{bai2024360} and OmniGS~\cite{li2025omnigs} extend 3DGS to panoramic modeling using plane-sphere coordinate dual mapping and equidistant cylindrical projection space transformations, respectively.
Our method maximally ensures compatibility with existing 3DGS pipelines while maintaining the fully explicit architecture and enabling efficient, high-quality rendering.
\section{Method}
\label{sec:method}
\setlength{\textfloatsep}{6pt}
\setlength{\floatsep}{6pt}
\setlength{\intextsep}{6pt}
We propose an embedded 3DGS framework based on the Kannala-Brandt fisheye projection model~\cite{kannala2006generic}. By deeply coupling the nonlinear projection mechanism into the 3D Gaussian splatting training process, we avoid the information loss caused by undistortion preprocessing while fully preserving the real-time advantages of rasterization rendering (Sec.~\ref{sec:fisheye modeling}). Additionally, we introduce a cross-view joint optimization strategy to further enhance the quality (Sec.~\ref{sec:cross-view}).

\subsection{Preliminary}

3DGS~\cite{kerbl20233d} represents a scene as a set of anisotropic 3D Gaussians:
$\mathcal{G} \;=\; \bigl\{ (\boldsymbol{\mu}_i,\;\boldsymbol{\Sigma}_i,\;\mathbf{c}_i,\;o_i)\bigr\}_{i=1}^N$,
where each Gaussian $i$ is parameterized by its positions ${\mu} \in \mathbb{R}^3$, rotations ${r} \in \mathbb{R}^4$, scale ${s} \in \mathbb{R}^3$, opacity ${o} \in \mathbb{R}$, and SH-based color $c$.
After projecting 3D Gaussians into 2D screen space and obtaining the 2D mean $\mu^{'}_i$ and covariance $\Sigma^{'}$, the image color $c(x)$ at pixel location $x$ is obtained through volume rendering formulation as follows:
\begin{equation}
    \alpha _{i}   =o_{i} \cdot \mathrm {exp} ( -\frac{1}{2}(x-\mu_{i}^{'})^{\mathrm{T}} \Sigma_{i}^{'-1}   (x-\mu_{i}^{'}) )
\end{equation}
\begin{equation}
    c(x)=\sum_{i=1}^{N} c_{i}\alpha _{i}\prod_{j=1}^{i-1}(1-\alpha _{j}) 
\end{equation}
The number $N$ of Gaussians that contribute to each pixel is determined through tile-based rasterization, which enables fast rendering of novel views.

\subsection{Fisheye Camera Model Formulation}
\label{sec:fisheye modeling}
\begin{figure}
    \centering
    \includegraphics[width=0.95\linewidth]{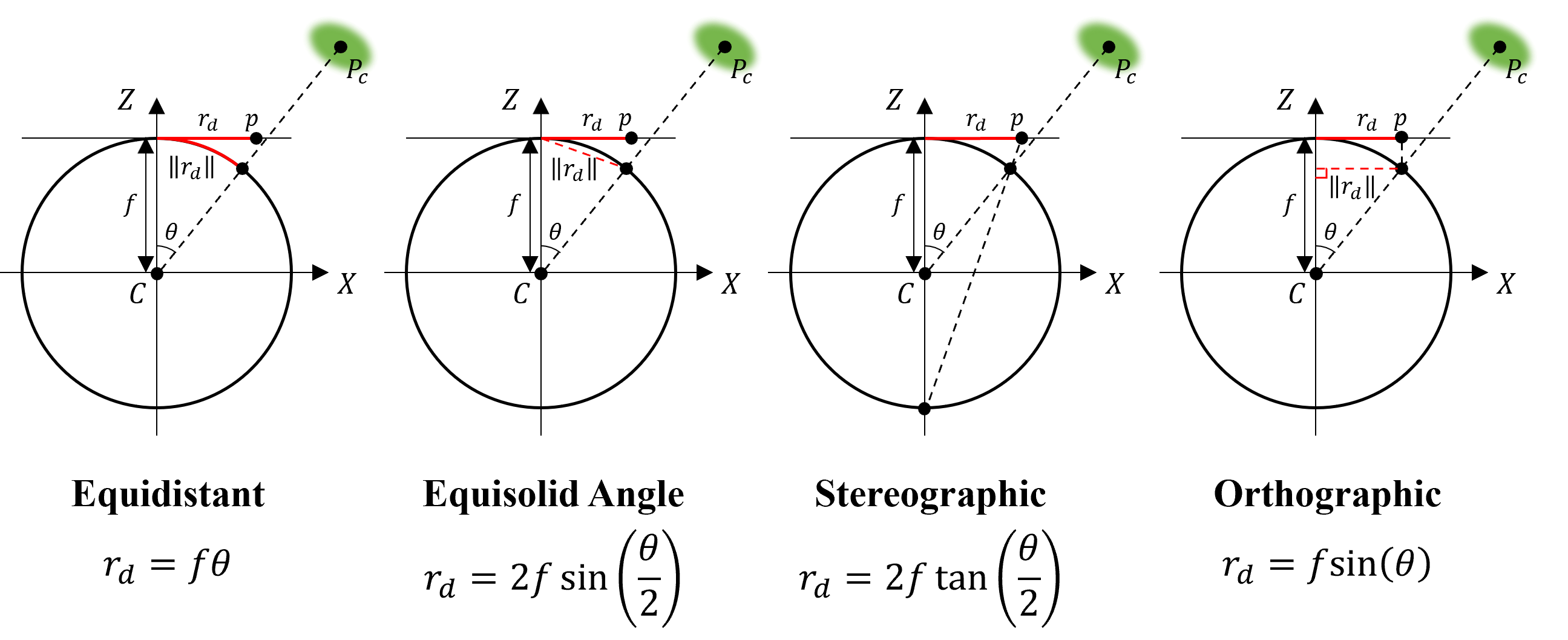}
    \caption{Common fisheye camera projection models. It is typically approximated as a unit sphere projection model. $P$ represents a Gaussian point in space, $C$ is the camera center, $\theta$ is the angle between the ray $PC$ and the camera optical axis, $f$ is the camera focal length, and $\mathbf r_d$ is the distance from the planar projection point $p$ to the image center.}
    \label{fig:fisheymodel}
\end{figure}

Traditional fisheye camera models use specific mathematical formulation, such as equidistant projection and equisolid angle projection, as shown in Fig.~\ref{fig:fisheymodel}. While these methods are computationally simple, they struggle to accurately describe the nonlinear distortions generated by complex optical systems. To establish a more general camera model, we use a polynomial expansion-based radial distortion approach, introducing four distortion parameters for a universal fisheye projection model.
\subsubsection{Polynomial Projection Model}
Given the 3D point coordinates in the camera coordinate system as $\mu_{cam}=(x_c, y_c, z_c)^T$, the corresponding angle of incidence of the ray can be defined as: 
\begin{equation}
\theta = arctan(\frac{\sqrt{x_c^2 + y_c^2}}{z_c})
\end{equation}
For different fisheye camera models, a Taylor expansion can be used for polynomial approximation, yielding the equivalent ray incidence angle~\cite{kannala2006generic}:
\begin{equation}
\theta_d = \theta + k_1 \theta^3 + k_2 \theta^5 + k_3 \theta^7 + k_4 \theta^9
\label{eq:poly_dist}
\end{equation}
Thus, the distance from the optical center to the projection point $r_d$ can be expressed as:
\begin{equation}
r_{dx} = f_x \cdot \theta_d , \quad r_{dy} = f_y \cdot \theta_d
\label{eq:poly_dist}
\end{equation}
where $f_x,f_y$ are the effective focal lengths in the x and y directions and $k_1,k_2,k_3,k_4$ are the distortion parameters obtained through calibration. This polynomial expansion can effectively characterize the complex radial distortion characteristics of various fisheye lenses.

In the case of considering the principal point offset and the non-orthogonality between axes, the pixel coordinates $\mu_p = (x_p, y_p)$ corresponding to a point $p$ in the camera coordinate system  can be expressed as:
\begin{equation}
x_p = c_x + \frac{r_{dx} \cdot x_c}{\sqrt{x_c^2 + y^2_c}}, \quad y_p = c_y + \frac{r_{dy} \cdot y_c}{\sqrt{x_c^2 + y^2_c}}
\end{equation}
where $(c_x,c_y)$ represents the pixel coordinates of the optical center.

\subsubsection{Forward Rendering and Gradient Calculation}
\label{subsubsection:f&b}
In 3DGS-based rendering framework, the differentiability of the forward projection process is crucial for achieving end-to-end optimization. To address the wide-angle imaging characteristics of fisheye cameras, we propose a Jacobian projection matrix computation method based on the Kannala-Brandt model~\cite{kannala2006generic}. For a spatial point $P_c$ in the camera coordinate system, we define its Euclidean distance to the camera center as $D=\|P_c\|_2=\sqrt{x_c^2+y_c^2+z_c^2}$, and its radial distance to the optical axis as $d=\sqrt{x_c^2+y_c^2}$. The projection Jacobian matrix $\mathbf{J_\theta} \in \mathbb{R}^{3\times3}$ is obtained by differentiating the pixel coordinates $\mu_p$ with respect to the camera space coordinates $\mu_{cam}$, and can be expressed as:
\begin{equation}
\label{Jacobian}
\begin{split}
\mathbf{J_\theta}=\frac{\partial \mu_p}{\partial \mu_{cam}}
  &= \theta_d' 
     \begin{bmatrix}
       f_x \dfrac{x_c^2 z_c}{D^2 d^2}  & 
       f_x \dfrac{x_c y_c z_c}{D^2 d^2} & 
       -f_x \dfrac{x_c}{D^2} \\
       f_y \dfrac{x_c y_c z_c}{D^2 d^2} & 
       f_y \dfrac{y_c^2 z_c}{D^2 d^2} & 
       -f_y \dfrac{y_c}{D^2} \\
       0 & 0 & 0
     \end{bmatrix} \\[8pt]
  &\quad +\,\theta_d 
     \begin{bmatrix}
       f_x \dfrac{y_c^2}{d^3}   & 
       -f_x \dfrac{x_c y_c}{d^3} & 
       0 \\
       -f_y \dfrac{x_c y_c}{d^3} & 
       f_y \dfrac{x_c^2}{d^3}   & 
       0 \\
       0 & 0 & 0
     \end{bmatrix}
\end{split}
\end{equation}
$\theta_d'$ is the first derivative of $\theta_d$ with respect to the incident
angle $\theta$ . The Eq.~\ref{Jacobian} fully characterizes the differential mapping from the 3D ellipsoid covariance $\boldsymbol{\Sigma}$ to 2D projected covariance $\boldsymbol{\Sigma}^{2D}$ in the fisheye case. The detailed rendering process and the backward gradient computation can be found in Supp.Algorithm~\ref{alg:forward} and Supp.Algorithm~\ref{alg:backward}.

\subsection{Cross-View Joint Optimization}
\label{sec:cross-view}
\begin{figure*}
    \centering
    \includegraphics[width=0.9\linewidth]{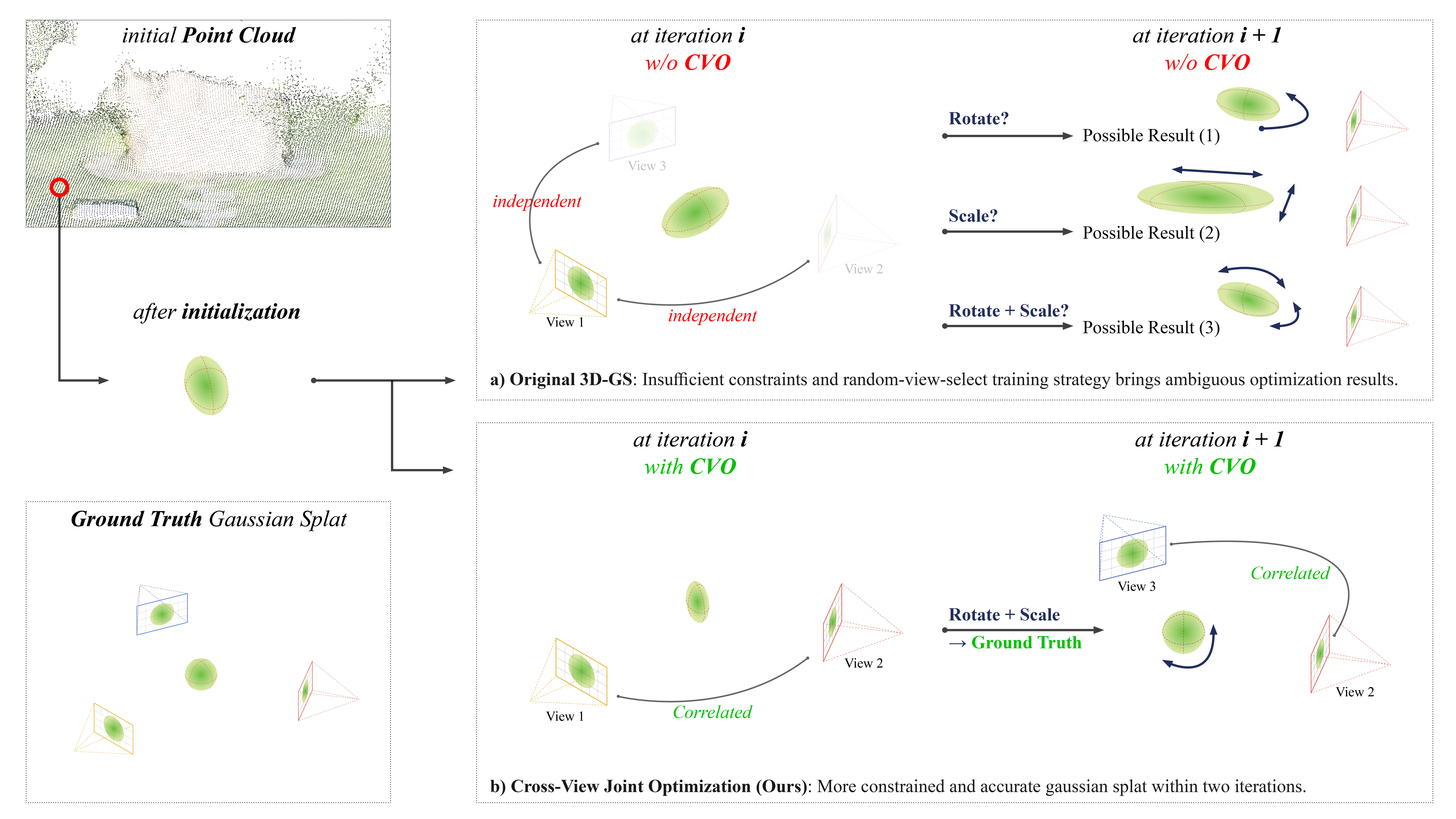}
    \caption{Illustration of the 3DGS single-view training paradigm and our proposed cross-view joint optimization strategy.}
    \label{fig:cross-view}
\end{figure*}

The original 3DGS training strategy randomly selects a single camera per iteration. While this randomness enhances viewpoint diversity, it introduces redundancy and ambiguity in the optimization process (Fig.~\ref{fig:cross-view}): optimizing a Gaussian in one viewpoint may drive transformations (e.g., translation, scaling, splitting) aligned with that view. However, abrupt viewpoint switches in subsequent iterations can trigger excessive Gaussian cloning, even in overlapping regions. Critically, independent optimization across iterations often yields inconsistent geometric or photometric parameters for the same Gaussian across views. As observed in experiments, even correctly model the fisheye camera, insufficient optimization leads to extreme Gaussian shapes and pronounced viewpoint anisotropy, especially at the edges of fisheye images, where distortion is greatest, causing noticeable floaters and flickering artifacts during head movements in VR and degrading immersion. Although MVGS~\cite{du2024mvgs} pioneers multi-view training, its random multi-camera sampling per iteration fails to leverage geometric correlations, resulting in limited quality gains beyond inflated Gaussian counts in some cases.

\begin{figure}
    \centering
    \includegraphics[width=\linewidth]{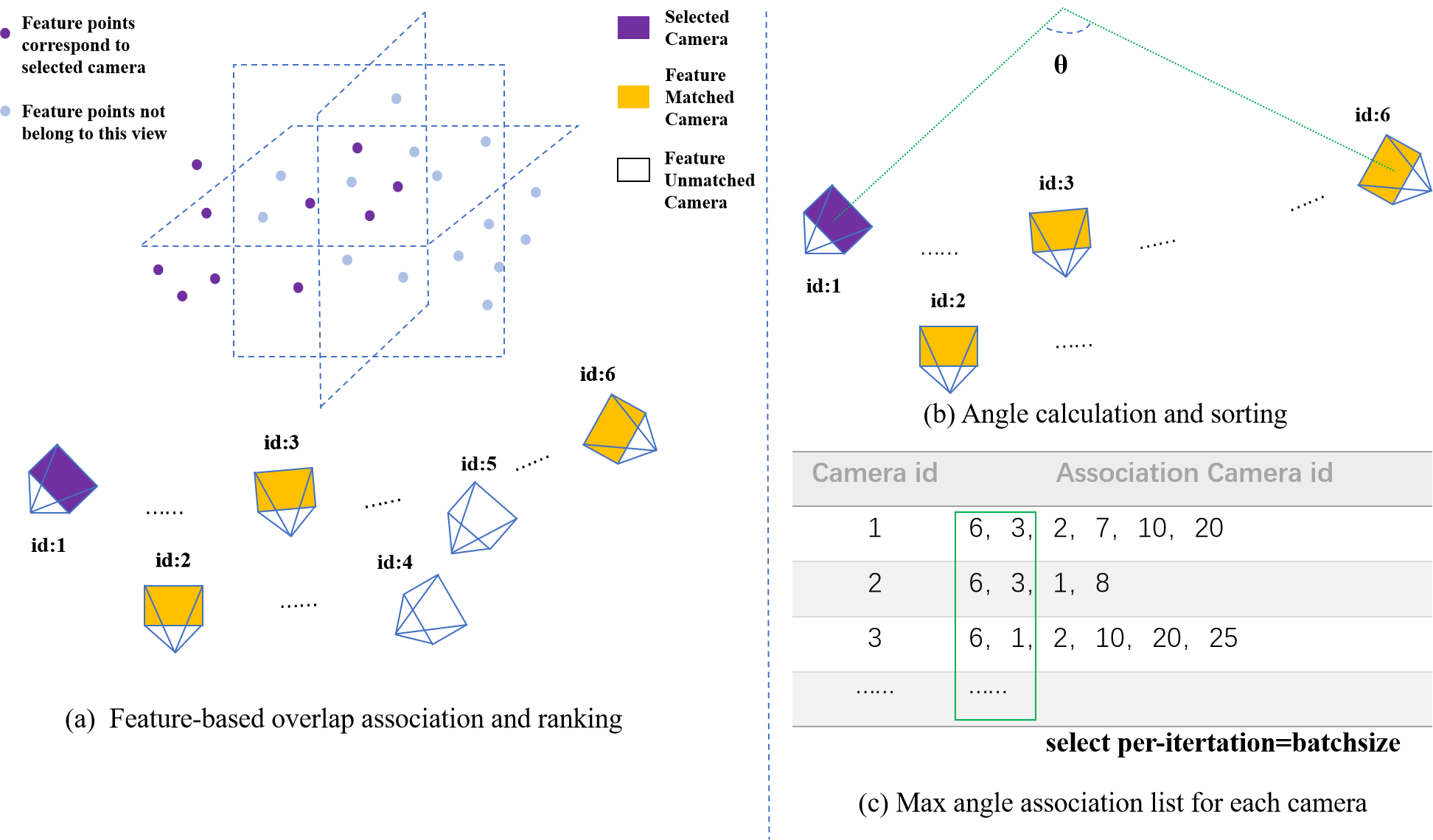}
    \caption{Schematic diagram of our camera association method.}
    \label{fig:view-selection}
\end{figure}

To address this issue, we propose a \textbf{Cross-View Joint Optimization (CVO)} framework that prioritizes geometrically overlapping viewpoints with maximal angular divergence. This ensures:

\noindent\textbf{1. Shared Gaussian Optimization:} Maximize the overlap of visible Gaussians across selected views during training.

\noindent\textbf{2. Multi-View Consistency: } Make global illumination more consistent and each Gaussian more 3D-consistent in shape and position, enabling better SH optimization.

The core challenge lies in defining ``overlap". While geometric overlap (viewing angle) is straightforward, for 3D reconstruction, we are more interested in describing the "information overlap" between cameras, which can be represented by feature points. Given that 3DGS initialization is based on SfM sparse point cloud (COLMAP~\cite{schoenberger2016sfm}), which inherently encodes feature correspondences across images, as shown in Fig.~\ref{fig:view-selection}, our implementation involves:

\noindent\textbf{Step 1: Camera Association.}
Construct a feature-overlap graph for each image, ranking neighboring cameras by shared SIFT feature counts \textit{($\rightarrow$ camera\_association graph)}. 

\noindent\textbf{Step 2: Angular Divergence Sorting.}
For each camera pair in the \textit{camera\_association graph}, compute their pose angular difference and sort them in descending order \textit{($\rightarrow$ max\_angle\_association graph)}.

During training, we iteratively sample a primary view and select its top-\textit{batchsize$-1$} correlated cameras from the graph. This strategy balances overlap and angular diversity, as formalized in Algorithm~\ref{alg:optimization}.

\begin{algorithm}[!h]
    \caption{Cross-View Joint Optimization}
    \label{alg:optimization}
    \begin{algorithmic}[1]
        \Require max\_angle\_association graph $\mathcal{G}_{\text{assoc}}$ (precomputed)
        \Require Batchsize $N$, total iterations $T$
        \State Initialize Gaussian parameters $\mathcal{G}$
        \For{$t \gets 1$ to $T$}
            \State Select a random camera $i$
            \If{$N > 1$}
                \State Retrieve top-$N-1$ cameras $\{j_1, \dots, j_N\}$ from $\mathcal{G}_{\text{assoc}}[i]$
                \State Initialize $L_{\text{total}} \gets 0$
                \ForAll{camera $j_k$ in $\{j_1, \dots, j_N\}$}
                    \State Render image $\hat{I}_k \gets \text{Render}(\mathcal{G}, j_k)$
                    \State $L_k \gets L_1(I_k,\hat{I}_k) + \text{SSIM}(I_k, \hat{I}_k)$ \Comment{Loss}
                    \State $L_{\text{total}} \gets L_{\text{total}} + L_k$  \Comment{Accumulate loss}
                \EndFor
                \State Backpropagate $L_{\text{total}}$ to update $\mathcal{G}$  \Comment{Optimization}
            \EndIf
        \EndFor
    \end{algorithmic}
\end{algorithm}
\section{Experiments}
\label{sec:experiments}

In this section, we provide details on the experiments to enable fair evaluation of 3DGS-based methods using fisheye camera data. We first introduce the baseline methods chosen for comparison (Sec.~\ref{baseline intro}), followed by the selected datasets and necessary pre-processing (Sec.~\ref{data pre}). Finally, we outline the hardware configuration used for training and the evaluation metrics adopted (Sec.~\ref{experiments setup}).

\subsection{Baselines for Comparisons}
\label{baseline intro}
Many follow-up works have improved performance of the original 3DGS~\cite{kerbl20233d}, but only a few have extended it to handle data from distorted cameras. Since the relevant improvements are well compatible with changes in the camera model, we choose to compare our method with the original 3DGS~\cite{kerbl20233d} and all existing open-source methods capable of handling fisheye cameras -Fisheye-GS~\cite{liao2024fisheye}, 3DGUT~\cite{wu20243dgut}, Self-Cali-GS (SC-GS)~\cite{deng2025self}.

It is worth noting that, for a more fair comparison, we evaluated the original 3DGS from two aspects: 1) Directly taking fisheye images as inputs and treating camera parameters as pinholes. 2) Take the undistorted images as inputs and use our rendering method to render fisheye images.

\subsection{Datasets Preparation}
\label{data pre}

To fully evaluate the effectiveness of our method, we conducted experiments using the two most commonly used public datasets and three large-scale indoor and outdoor scenarios.
\vspace{-10pt}
\paragraph{FisheyeNeRF~\cite{jeong2021self} } is captured with a fisheye camera. It contains 6 object-centric small-scale scenes with a resolution of 4240×2384. We run COLMAP~\cite{schoenberger2016sfm} using the camera intrinsic parameters provided by the author as prior to obtain the point cloud. The inputs for the other two methods are the same, except for Fisheye-GS, which requires processing into an ideal equidistant fisheye camera.
\vspace{-10pt}
\paragraph{Scannet++~\cite{yeshwanth2023scannet++}}  is a middle-scale dataset containing over 1,000 3D indoor scenes captured by fisheye cameras. We use the resized images with a resolution of 1752×1168, selecting the same six scenes as in Fisheye-GS~\cite{liao2024fisheye} and 3DGUT~\cite{wu20243dgut}. Notably, for Fisheye-GS~\cite{liao2024fisheye}, we convert the images to an equidistant fisheye camera model as required for successful training, while we use the original calibration results provided for 3DGUT~\cite{wu20243dgut} and DirectFisheye-GS (Ours).
\vspace{-10pt}
\paragraph{Den-SOFT~\cite{yu2024soft}} includes 9 large-scale real-world scenes for VR/AR applications. Unlike previous volumetric video datasets, it was captured using 46 motion cameras at very high density and resolution (5568×4176) in an inside-looking-out manner. Among them, the \textit{Ruziniu} and \textit{Coffee} scenes provide fisheye data. We evaluated the above-mentioned methods at a resolution of 1.6k to truly validate the feasibility of our approach for user-friendly real-world applications.

\subsection{Experiments Setup}
\label{experiments setup}
\paragraph{Training}All experiments are done on a single NVIDIA A100 80GB GPU. We adopt the default hyperparameters for all baselines to ensure a fair comparison. The batchsize is set to 2 in our method.
\vspace{-10pt}
\paragraph{Metrics}We used the three most common metrics in NVS tasks, peak signal-to-noise ratio (PSNR), learned perceptual image patch similarity (LPIPS-VGG), and structural similarity (SSIM), for quantitative evaluation of the algorithm's performance.
\section{Results and Analysis}
\label{sec:results}

\subsection{Qualitative and Quantitative Results}
\begin{figure*}
    \centering
    \includegraphics[width=\textwidth]{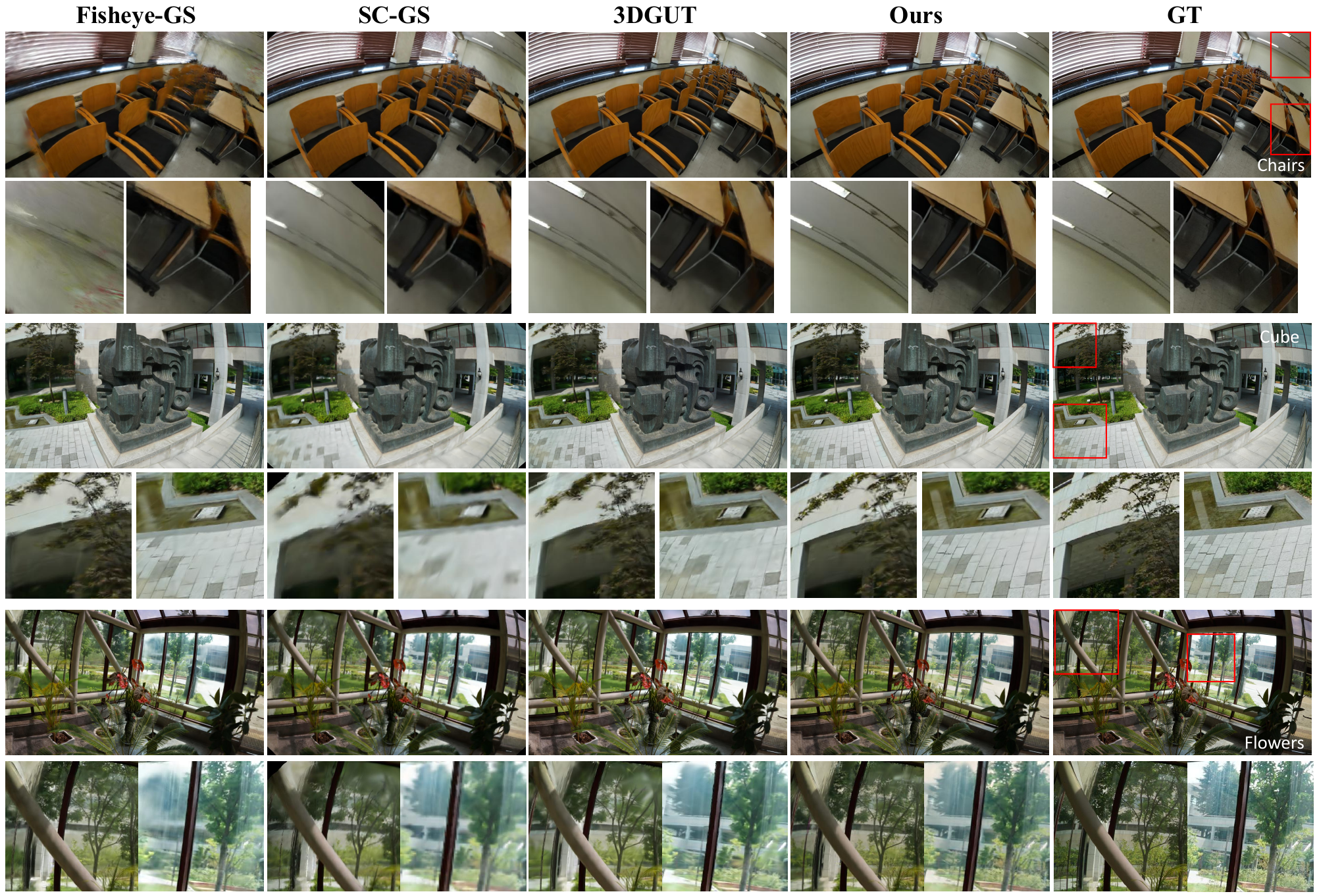}
    \caption{Qualitative comparison of our results against the baselines on the FisheyeNeRF~\cite{jeong2021self} dataset. Ours has more details, better texture, and fewer floaters.}
    \label{fig:fisheyenerf}
\end{figure*}

\begin{table*}[h]
\centering
\caption{Per-scene results of our method and related state-of-art works on the FisheyeNeRF~\cite{jeong2021self} Dataset (Train View). * shows the results of undistorted images as inputs and fisheye rendered images as output. \textbf{** Fisheye-GS still need to pre-process data before training.}}
\label{tab:fisheyenerf}
\resizebox{\textwidth}{!}{%
\begin{tabular}{lccccccccccccccccccccc}
\toprule
\multicolumn{1}{c}{}                                       & \multicolumn{3}{c}{Globe}                                                                         & \multicolumn{3}{c}{Rock}                                                                          & \multicolumn{3}{c}{Cube}                                                                          & \multicolumn{3}{c}{Flowers}                                                                       & \multicolumn{3}{c}{Heart}                                                                         & \multicolumn{3}{c}{Chairs}                                                                        & \multicolumn{3}{c}{\textbf{Average}}                                                                                         \\
\multicolumn{1}{c}{\multirow{-2}{*}{FisheyeNeRF}} & SSIM↑                          & PSNR↑                           & LPIPS↓                         & SSIM↑                          & PSNR↑                           & LPIPS↓                         & SSIM↑                          & PSNR↑                           & LPIPS↓                         & SSIM↑                          & PSNR↑                           & LPIPS↓                         & SSIM↑                          & PSNR↑                           & LPIPS↓                         & SSIM↑                          & PSNR↑                           & LPIPS↓                         & \textbf{SSIM↑}                          & \textbf{PSNR↑}                           & \textbf{LPIPS↓}                         \\ \midrule
\textbf{3DGS}                                              & 0.6355                         & 19.8407                         & 0.5273                         & 0.5306                         & 18.0784                         & 0.5122                         & 0.6491                         & 19.3742                         & 0.5388                         & 0.5412                         & 18.1212                         & 0.5362                         & 0.6842                         & 19.8638                         & 0.5045                         & 0.6338                         & 17.7938                         & 0.5179                         & \textbf{0.6124}                         & \textbf{18.8454}                                & \textbf{0.5228}                         \\
\textbf{3DGS*}                                             & 0.8587                         & 26.9129                         & 0.2223                         & 0.8108                         & 26.6751                         & 0.2129                         & 0.8356                         & 26.4027                         & 0.2502                         & 0.8187                         & 24.5977                         & 0.2140                         & 0.8163                         & 25.6968                         & 0.2666                         & 0.8036                         & 22.9437                         & 0.2924                         & \textbf{0.8240}                         & \textbf{25.5382}                                & \textbf{0.2431}                         \\ \hline
\textbf{Fisheye-GS**}                                       & \cellcolor[HTML]{FFCB2F}0.8593 & \cellcolor[HTML]{FFCB2F}26.9145 & \cellcolor[HTML]{FFFE65}0.2245                         & \cellcolor[HTML]{FFCB2F}0.8230 & \cellcolor[HTML]{FFCB2F}26.8117 & \cellcolor[HTML]{FFFE65}0.2242                         & \cellcolor[HTML]{FFCB2F}0.8384 & \cellcolor[HTML]{FFFE65}26.9748                         & \cellcolor[HTML]{FFFE65}0.2604                         & \cellcolor[HTML]{FFCB2F}0.8022 & \cellcolor[HTML]{FFCB2F}24.4500 & \cellcolor[HTML]{FFCB2F}0.2285 & \cellcolor[HTML]{FFCB2F}0.8223 & \cellcolor[HTML]{FFFE65}24.9219 & \cellcolor[HTML]{FFFE65}0.2888 & 0.7643                         & 20.9969                         & 0.3681                         & \textbf{0.8183}                         & \textbf{25.1783}                         & \cellcolor[HTML]{FFFE65}\textbf{0.2658} \\
\textbf{3DGUT}                                             & 0.8200                         & 26.0860                         & 0.3250                         & 0.7730                         & 25.4410                         & 0.2970                         & 0.8050                         & 26.0730                         & 0.3530                         & 0.7340                         & 23.3250                         & 0.3370                         & 0.7890                         & 24.5390                         & 0.3840                         & \cellcolor[HTML]{FFFE65}0.8910 & \cellcolor[HTML]{FFFE65}26.2380 & \cellcolor[HTML]{FFFE65}0.2780 & \cellcolor[HTML]{FFFE65}\textbf{0.8020} & \cellcolor[HTML]{FFFE65}\textbf{25.2837} & \textbf{0.3290}                         \\
\textbf{Self-Cali-GS}                                              & 0.7428                         & 24.6431                         & 0.4628                         & 0.7162                         & 24.4566                         & 0.4955                         & 0.7638                         & 24.9037                         & 0.4702                         & 0.6458                         & 22.1389                         & 0.5398                         & 0.8023                         & 24.6079                         & 0.3839                         & \cellcolor[HTML]{FFFFFF}0.8048 & \cellcolor[HTML]{FFFFFF}23.3000 & \cellcolor[HTML]{FFFFFF}0.3522 & \textbf{0.7460}                         & \textbf{24.0084}                         & \textbf{0.4507}                         \\
\textbf{Ours}                                              & \cellcolor[HTML]{FFFE65}0.8351 & \cellcolor[HTML]{FFFE65}26.6411 & \cellcolor[HTML]{FFCB2F}0.2111 & \cellcolor[HTML]{FFFE65}0.8066 & \cellcolor[HTML]{FFFE65}26.6411 & \cellcolor[HTML]{FFC702}0.2228 & \cellcolor[HTML]{FFFE65}0.8259 & \cellcolor[HTML]{FFCB2F}27.0780 & \cellcolor[HTML]{FFCB2F}0.2491 & \cellcolor[HTML]{FFFE65}0.7802 & \cellcolor[HTML]{FFFE65}24.1444 & \cellcolor[HTML]{FFFE65}0.2335 & \cellcolor[HTML]{FFFE65}0.8178 & \cellcolor[HTML]{FFCB2F}25.5095 & \cellcolor[HTML]{FFCB2F}0.2717 & \cellcolor[HTML]{FFCB2F}0.9049 & \cellcolor[HTML]{FFCB2F}27.4988 & \cellcolor[HTML]{FFCB2F}0.1890 & \cellcolor[HTML]{FFCB2F}\textbf{0.8284} & \cellcolor[HTML]{FFCB2F}\textbf{26.2522} & \cellcolor[HTML]{FFCB2F}\textbf{0.2295} \\ \bottomrule
\end{tabular}%
}
\end{table*}

As shown in Table~\ref{tab:fisheyenerf},~\ref{tab:scannet++},~\ref{tab:Den-SOFT} and Supp.Table~\ref{tab:fisheyenerf_test}, our method consistently all baselines across various datasets, metrics, and scene configurations—including both training and test views, indoor and outdoor environments, and small- to large-scale scenes.
The original 3DGS cannot handle fisheye images, scores lowest in every case. However, when fisheye images are undistorted and used for 3DGS training, followed by fisheye rendering, our method achieves superior or comparable results in all cases, demonstrating the correctness of our camera modeling and the fact that undistortion inevitably loses some scene information.

Fig.~\ref{fig:fisheyenerf} and Supp.Fig.~\ref{fig:den-soft} illustrate the qualitative comparisons of baseline methods with ours. Fisheye-GS’s simplistic model yields steep quality degradation toward the image edges. SC-GS, which learns a deformation field via a neural network, gains stability but introduces blurriness due to the smoothness and continuity of the neural network, and still loses edge information. 3DGUT achieves performance comparable to ours overall, but edge regions suffer from blurred artifacts due to Gaussians whose gradients are not fully aligned with the camera model and the lack of cross-view joint constraints; its central regions also exhibit discontinuous lighting artifacts from over-emphasized illumination modeling. These limitations become more severe in unbounded outdoor scenes—where lighting varies dramatically and fine detail abounds—so on an outdoor scene like Ruziniu, our method significantly outperforms others. Our results can be viewed using the original SIBR\_Viewer, and we achieve results comparable to training 3DGS with undistorted images. More in-depth comparisons and analysis can be seen in supplementary materials (Sec.~\ref{sec:more results}).

\begin{table*}[]
\centering
\caption{Per-scene evaluation results of our method and related state-of-art works on the Scannet++~\cite{yeshwanth2023scannet++} Dataset (Test View).}
\label{tab:scannet++}
\resizebox{\textwidth}{!}{%
\begin{tabular}{lccccccccccccccccccccc}
\toprule
\multicolumn{1}{c}{}                                     & \multicolumn{3}{c}{\textbf{0a5c013435}}                                                           & \multicolumn{3}{c}{\textbf{8d563fc2cc}}                                                           & \multicolumn{3}{c}{\textbf{bb87c292ad}}                                                           & \multicolumn{3}{c}{\textbf{d415cc449b}}                                                           & \multicolumn{3}{c}{\textbf{e8ea9b4da8}}                                                           & \multicolumn{3}{c}{\textbf{fe1733741f}}                                                           & \multicolumn{3}{c}{\textbf{Average}}                                                              \\
\multicolumn{1}{c}{\multirow{-2}{*}{\textbf{Scannet++}}} & \textbf{SSIM↑}                 & \textbf{PSNR↑}                  & \textbf{LPIPS↓}                & \textbf{SSIM↑}                 & \textbf{PSNR↑}                  & \textbf{LPIPS↓}                & \textbf{SSIM↑}                 & \textbf{PSNR↑}                  & \textbf{LPIPS↓}                & \textbf{SSIM↑}                 & \textbf{PSNR↑}                  & \textbf{LPIPS↓}                & \textbf{SSIM↑}                 & \textbf{PSNR↑}                  & \textbf{LPIPS↓}                & \textbf{SSIM↑}                 & \textbf{PSNR↑}                  & \textbf{LPIPS↓}                & \textbf{SSIM↑}                 & \textbf{PSNR↑}                  & \textbf{LPIPS↓}                \\ \midrule
\textbf{3DGS}                                            & 0.8063                               & 18.3261                                & 0.3430                               & 0.7782                               & 15.6411                                & 0.3899                               & 0.8274                               & 19.8864                                & 0.3522                               & 0.5999                               & 15.5700                                & 0.5401                               & 0.8613                               & 20.8932                                & 0.3445                               & 0.6791                               & 14.8815                                & 0.4825                               & 0.7587                                & 17.5331                                 & 0.4087                                \\
\textbf{3DGS*}                                           & 0.8754                         & 24.5032                         & 0.1918                         & 0.8903                         & 24.9394                         & 0.1894                         & 0.9148                         & 28.6049                         & 0.2044                         & 0.7693                         & 23.8866                         & 0.2780                         & 0.9271                         & 29.6452                         & 0.2127                         & 0.8137                         & 23.6730                         & 0.2573                         & 0.8651                         & 25.8754                         & 0.2223                         \\ \hline
\textbf{Fisheye-GS**}                                     & 0.8940                         & 26.4200                         & \cellcolor[HTML]{FFFE65}0.1770 & \cellcolor[HTML]{FFCC67}0.9050 & \cellcolor[HTML]{FFCC67}26.1500 & \cellcolor[HTML]{FFFE65}0.1770 & \cellcolor[HTML]{FFFE65}0.9330 & \cellcolor[HTML]{FFCC67}30.6300 & \cellcolor[HTML]{FFFE65}0.1790 & \cellcolor[HTML]{FFFE65}0.8550 & 27.1200                         & \cellcolor[HTML]{FCFF2F}0.2230 & 0.9460                         & 31.8900                         & 0.1910                         & \cellcolor[HTML]{FFCC67}0.8490 & \cellcolor[HTML]{FFFE65}24.9500 & \cellcolor[HTML]{FFFE65}0.2230 & 0.8970                         & 27.8600                         & \cellcolor[HTML]{FFFE65}0.1950 \\
\textbf{3DGUT}                                           & \cellcolor[HTML]{FFCC67}0.9300 & \cellcolor[HTML]{FFCC67}29.7930 & 0.2290                         & \cellcolor[HTML]{FFFE65}0.9010 & 25.7450                         & 0.2520                         & \cellcolor[HTML]{FFCC67}0.9340 & 30.4110                         & 0.2470                         & \cellcolor[HTML]{FFCC67}0.8650 & \cellcolor[HTML]{FFCC67}27.4000 & 0.2630                         & \cellcolor[HTML]{FFCC67}0.9510 & \cellcolor[HTML]{FFFE65}32.2130 & 0.2570                         & \cellcolor[HTML]{FFFE65}0.8440 & 24.9450                         & 0.2900                         & \cellcolor[HTML]{FFCC67}0.9042 & \cellcolor[HTML]{FFCC67}28.4178 & 0.2563                         \\
\textbf{Self-Cali-GS}                                            & 0.8549                         & 22.7834                         & 0.1985                         & 0.8422                         & 21.1700                         & 0.2316                         & 0.8747                         & 25.6795                         & 0.2055                         & 0.7259                         & 21.9833                         & 0.3383                         & 0.9081                         & 28.1148                         & \cellcolor[HTML]{FFCC67}0.1481 & 0.7447                         & 20.0664                         & 0.3431                         & 0.8251                         & 23.2996                         & 0.2442                         \\
\textbf{Ours}                                            & \cellcolor[HTML]{FFFE65}0.9191 & \cellcolor[HTML]{FFFE65}28.7369 & \cellcolor[HTML]{FFCC67}0.1511 & 0.8984                         & \cellcolor[HTML]{FFFE65}25.9518 & \cellcolor[HTML]{FFCC67}0.1715 & 0.9296                         & \cellcolor[HTML]{FFFE65}30.6031 & \cellcolor[HTML]{FFCC67}0.1783 & 0.8499                         & \cellcolor[HTML]{FFFE65}27.1832 & \cellcolor[HTML]{FFCC67}0.2127 & \cellcolor[HTML]{FFFE65}0.9473 & \cellcolor[HTML]{FFCC67}32.7372 & \cellcolor[HTML]{FFFE65}0.1849 & 0.8390                         & \cellcolor[HTML]{FFCC67}24.9978 & \cellcolor[HTML]{FFCC67}0.2194 & \cellcolor[HTML]{FFFE65}0.8972 & \cellcolor[HTML]{FFFE65}28.3683 & \cellcolor[HTML]{FFCC67}0.1863 \\ \bottomrule
\end{tabular}%
}
\end{table*}

\begin{table}[]
\caption{Performance of our method and four baselines on Den-SOFT~\cite{yu2024soft} dataset (Train View \& Test View).}
\label{tab:Den-SOFT}
\resizebox{\columnwidth}{!}{%
\begin{tabular}{lcccccc}
\toprule
\multicolumn{1}{c}{}                                      & \multicolumn{3}{c}{\textbf{Ruziniu}}                                                              & \multicolumn{3}{c}{\textbf{Coffee}}                                                               \\
\multicolumn{1}{c}{\multirow{-2}{*}{\textbf{Train View}}} & \textbf{SSIM↑}                 & \textbf{PSNR↑}                  & \textbf{LPIPS↓}                & \textbf{SSIM↑}                 & \textbf{PSNR↑}                  & \textbf{LPIPS↓}                \\ \midrule
\textbf{3DGS*}                                            & \cellcolor[HTML]{FFFE65}0.7518 & 20.0220                         & \cellcolor[HTML]{FFFE65}0.2697 & 0.8826                         & 24.7996                         & \cellcolor[HTML]{FFFE65}0.2416 \\
\textbf{Fisheye-GS**}                                        & 0.5965                         & 20.8819                         & 0.4447                         & 0.8737                         & 25.6869                         & 0.2661                         \\
\textbf{3DGUT}                                            & 0.7500                         & \cellcolor[HTML]{FFFE65}22.2100 & 0.3400                         & 0.8830                         & \cellcolor[HTML]{FFFE65}25.7450 & 0.3310                         \\
\textbf{Self-Cali-GS}                                             & 0.5224                         & 19.7540                         & 0.5969                         & \cellcolor[HTML]{FFFE65}0.8873 & 22.9550                         & 0.2460                         \\
\textbf{Ours}                                             & \cellcolor[HTML]{FFCC67}0.8225 & \cellcolor[HTML]{FFCC67}24.0154 & \cellcolor[HTML]{FFCC67}0.2140 & \cellcolor[HTML]{FFCC67}0.9105 & \cellcolor[HTML]{FFCC67}27.3528 & \cellcolor[HTML]{FFCC67}0.2181 \\ \bottomrule
\multicolumn{1}{c}{\textbf{Test View}}                    & \textbf{SSIM↑}                 & \textbf{PSNR↑}                  & \textbf{LPIPS↓}                & \textbf{SSIM↑}                 & \textbf{PSNR↑}                  & \textbf{LPIPS↓}                \\ \midrule
\textbf{3DGS*}                                            & \cellcolor[HTML]{FFFE65}0.7403 & 19.6615                         & \cellcolor[HTML]{FFFE65}0.2754 & 0.8740                         & 24.0067                         & \cellcolor[HTML]{FFFE65}0.2488 \\
\textbf{Fisheye-GS**}                                        & 0.5843                         & \cellcolor[HTML]{FFFE65}19.9809 & 0.4523                         & 0.8664                         & 24.6196                         & 0.2731                         \\
\textbf{3DGUT}                                            & 0.5230                         & 19.8780                         & 0.4380                         & \cellcolor[HTML]{FFFE65}0.8770 & \cellcolor[HTML]{FFFE65}24.9090 & 0.3390                         \\
\textbf{Self-Cali-GS}                                             & 0.5072                         & 19.5411                         & 0.6024                         & 0.8629                         & 21.4954                         & 0.2729                         \\
\textbf{Ours}                                             & \cellcolor[HTML]{FFCC67}0.8006 & \cellcolor[HTML]{FFCC67}22.7404 & \cellcolor[HTML]{FFCC67}0.2222 & \cellcolor[HTML]{FFCC67}0.8815 & \cellcolor[HTML]{FFCC67}24.9305 & \cellcolor[HTML]{FFCC67}0.2299 \\ \bottomrule
\end{tabular}%
}
\end{table}

\begin{table}[h]
\centering
\small 
\setlength{\tabcolsep}{4pt} 
\caption{PSNR comparisons of the image-boundary region on Den-SOFT~\cite{yu2024soft} dataset.}
\label{tab:edge_region}
\begin{tabular*}{\columnwidth}{l @{\extracolsep{\fill}} ccc} 
\toprule
    \textbf{Scene} & \textbf{w/o CVO} & \textbf{w/ CVO} & \textbf{3DGUT} \\
    \midrule
    Ruziniu & 23.0033 & \textbf{23.7625} & 22.2242 \\
    Coffee  & 26.2492 & \textbf{27.3819} & 25.8105 \\
    \bottomrule
\end{tabular*}
\end{table}

\subsection{Ablation Study}
\begin{figure}
    \centering
    \includegraphics[width=\linewidth]{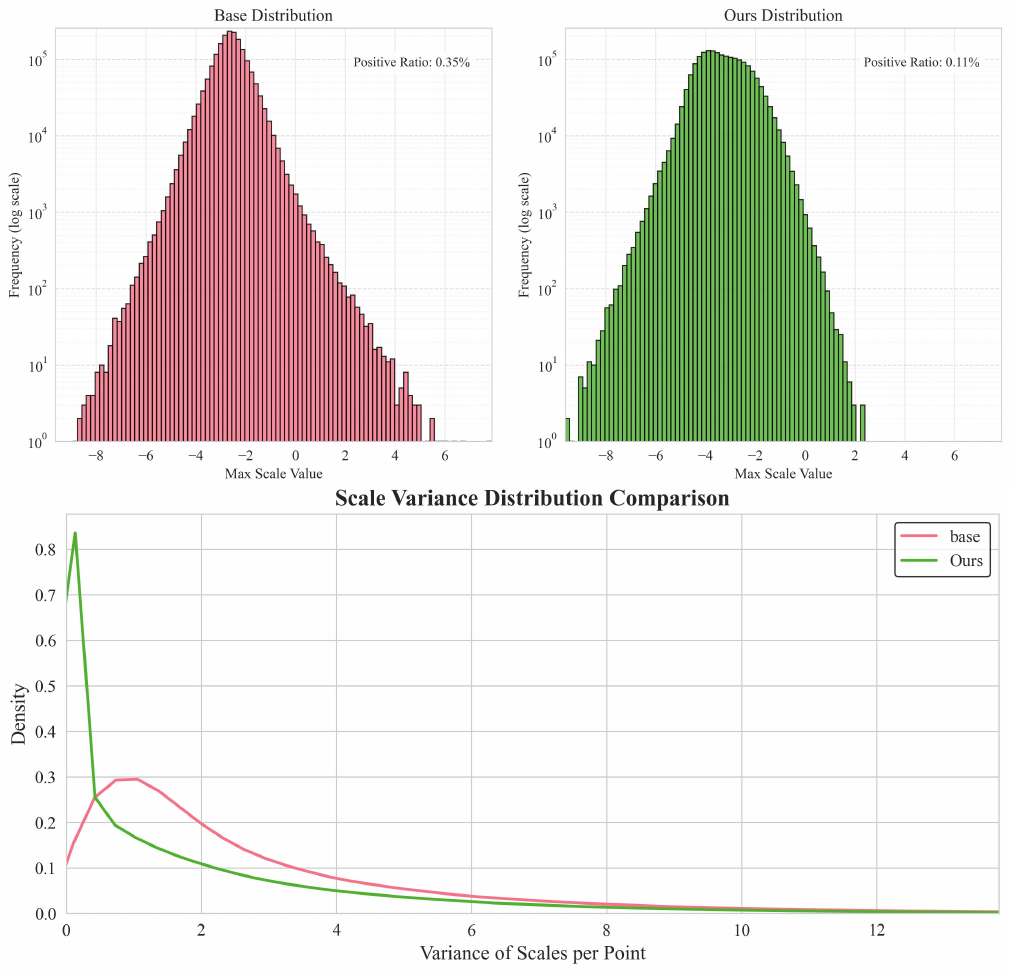}
    \caption{Gaussian's scales statistics for scene: {\itshape FisheyeNeRF-Chairs}. \textbf{Base:} original 3DGS with fisheye model. \textbf{Ours:} fisheye + CVO.}
    \label{fig:scales statistic}
\end{figure}

\paragraph{CVO performance under fisheye inputs.}
As shown in Table ~\ref{tab:CVO-fisheye} and Supp.Fig.~\ref{fig:ablation-study}(a), adding CVO markedly reduces oversized Gaussians at the edges of fisheye images, yielding much less floaters. The center of the image also exhibits enhanced details due to CVO. To be noted that, the ``random‑select" strategy in Table~\ref{tab:CVO-fisheye} matches MVGS’s~\cite{du2024mvgs} per‑iteration sampling, but MVGS also adds orthogonal improvement (e.g., densification). To isolate our contribution, we labeled it “random‑select”. We further visualize the distribution of Gaussian scales for the same scene \textit{w} or \textit{w/o} CVO in Fig.~\ref{fig:scales statistic}: after applying CVO, Gaussian sizes across the scene become far more uniform, and extreme shapes are greatly diminished. We also demonstrated the impact of different views selection strategies on the results. As shown in Table~\ref{tab:view-selection} and Supp.Fig.~\ref{fig:ablation-study}(c), by selecting image-pairs in each iteration, our method achieves higher quality and fewer points compared to random selection.

We further compared our method's performance with the SOTA method at the edges of fisheye images, as shown in Table~\ref{tab:edge_region}. CVO improves PSNR at the boundary (beyond 60° FOV) and outperforms 3DGUT, reinforcing the effectiveness of the fisheye model and the optimization of CVO for Gaussians at image edges.
\begin{table}[]
\caption{The performance of Cross-View Joint Optimization (CVO) under native fisheye inputs.}
\label{tab:CVO-fisheye}
\resizebox{\columnwidth}{!}{%
\begin{tabular}{lcccccc}
\toprule
                  & \multicolumn{3}{c}{\textbf{Ruziniu}}                                 & \multicolumn{3}{c}{\textbf{Coffee}}                                  \\
\textbf{Method}   & \textbf{SSIM↑} & \textbf{PSNR↑} & \textbf{LPIPS↓} & \textbf{SSIM↑} & \textbf{PSNR↑} & \textbf{LPIPS↓} \\ \midrule
Fisheye model           & 0.8172        & 23.7717       & 0.2230                     & 0.9001              & 27.1948              & 0.2223               \\
Fisheye model + CVO(Ours) & 0.8225        & 24.0154       & 0.2140                    & 0.9105              & 27.3528              & 0.2181               \\ \bottomrule
\end{tabular}%
}
\end{table}

\begin{table}[]
\centering
\caption{The results of adopting different views selection strategies. ``Random-select" refers to training strategy used in MVGS~\cite{du2024mvgs}.}
\label{tab:view-selection}
\resizebox{\columnwidth}{!}{%
\begin{tabular}{lcccc}
\toprule
\multicolumn{1}{l}{}                              & \multicolumn{4}{c}{\textbf{Self-captured Dataset: Ruziniu}}                                     \\
\textbf{Method}                                   & \textbf{SSIM↑} & \textbf{PSNR↑} & \textbf{LPIPS↓} & \textbf{Num.Points} \\ \midrule
Fisheye model                            & 0.8172        & 23.7717       & 0.2230         & 6330482             \\
Fisheye model + batchsize (random select) & 0.8230        & 23.5864       & 0.2174         & 6872364             \\
Fisheye model + cross-view (ours)        & 0.8225        & 24.0154       & 0.2140         & 6668098             \\ \bottomrule
\end{tabular}%
}
\end{table}

\paragraph{CVO performance under pinhole inputs.}

\begin{table}[]
\caption{Cross-view joint optimization (CVO) performance under pinhole inputs.}
\label{tab:CVO-pinhole}
\resizebox{\columnwidth}{!}{%
\begin{tabular}{lcccccc}
\toprule
                           & \multicolumn{3}{c}{\textbf{Square2}}                                 & \multicolumn{3}{c}{\textbf{Coffee}}                                                                                                                  \\
\textbf{Method}            & \textbf{SSIM↑} & \textbf{PSNR↑} & \textbf{LPIPS↓} & \multicolumn{1}{c}{\textbf{SSIM↑}} & \multicolumn{1}{c}{\textbf{PSNR↑}} & \multicolumn{1}{c}{\textbf{LPIPS↓}}  \\ \midrule
3DGS & 0.9354        & 30.5345       & 0.1505                      & 0.9013                                  & 27.2348                                   & 0.2237                                        \\
3DGS + random select  & 0.9355        & 30.6498       & 0.1491                      & 0.8936                                  & 26.2880                                   & 0.2370                                        \\
3DGS + CVO         & \textbf{0.9403}      & \textbf{30.9500}       & \textbf{0.1420}                     & \textbf{0.9033}                                  & \textbf{27.5138}                                  & \textbf{0.2172}                                      \\ \bottomrule
\end{tabular}%
}
\end{table}

This strategy can also be seamlessly integrated into the pinhole camera–based GS pipeline. Results in Table~\ref{tab:CVO-pinhole} and Supp.~Fig.~\ref{fig:ablation-study}(b) show that adding CVO to the vanilla 3DGS consistently improves reconstruction quality over the baseline and outperforms random-selection. We further validate this trend on additional public datasets (see Supp.Table~\ref{tab:CVO-tnt}).
\section{Conclusion}
\label{sec:conclusion}
In this work, we propose DirectFisheye-GS, an extension of 3D Gaussian Splatting that directly ingests fisheye images without undistortion. By mathematically modeling the fisheye projection and deriving its gradients, we seamlessly integrate fisheye support into the existing CUDA framework. Moreover, our novel cross-view joint optimization replaces the original per-iteration random view selection, enforcing consistent geometric and photometric constraints of a Gaussian across views. This not only enhances reconstruction quality—particularly at the highly distorted image edges—but also applies equally to pinhole-camera pipelines.
Extensive experiments on public benchmarks and large-scale real-world VR datasets demonstrate that DirectFisheye-GS matches or exceeds state-of-the-art methods. In future work, we will explore how advances in pinhole-camera–based techniques can be transferred to further boost fisheye reconstruction performance.
\section*{Acknowledgment}
{\small
This work was supported by the National Natural Science Foundation of China (NSFC) under Grant No. 62422110 and 62171253, and the Tsinghua University-Fuzhou Joint Institute for Data Technology (No. JIDT2024003). The authors also acknowledge the support provided by the collaborative research project between Tsinghua University and JD.com, Inc. Furthermore, we are grateful to Shi Pan for his assistance in visual asset optimization and supplementary video editing.}
{
    \small
    \bibliographystyle{ieeenat_fullname}
    \bibliography{main}
}

\clearpage
\setcounter{page}{1}
\maketitlesupplementary

\appendix
\section{Overview}
\label{sec:rationale}
Within the supplementary material, we provide:
\begin{itemize}
\item In Section~\ref{sec:reiterate}, we further elaborate on the motivation and intent behind our current work;
\item Section~\ref{sec:algorithm} provides the pseudocode for the KB-fisheye model-based forward rendering and backward gradient propagation algorithm mentioned in Section~\ref{subsubsection:f&b} of the main text; We further provide an intuitive mathematical analysis explaining why regions with strong distortion tend to produce extremely anisotropic Gaussians mentioned in Section~\ref{sec:cross-view}.
\item In Section~\ref{sec:more results}, we present a comparative analysis of Fisheye-GS, 3DGUT, SC-GS, and our method on the Den-SOFT dataset, along with a more detailed evaluation of the three approaches;
\item In Sections~\ref{sec:larger batch-size} and Section~\ref{sec:computational cost}, we discuss the performance of our method with larger batch sizes and its time efficiency, respectively;
\item In Section~\ref{sec:limitations}, we address the limitations of this work.
\end{itemize}
Additionally, we include a supplementary video that demonstrates the workflow and related visualizations, featuring more qualitative comparisons, demonstrations in VR, and more. Please see our video for details.

\section{Reiterate Our Contributions and Insights:}
\label{sec:reiterate}
\textbf{1. KB Fisheye Model Integration:} We are the first to introduce KB-model-based rasterization and back-propagation into 3DGS without replacing the efficient splatting with relatively heavy ray-tracing.
This eliminates the need for fisheye image pinhole/equivalent undistortion (3DGS/Fisheye-GS), which will loss boundary pixels and sacrifice details by simple bilinear-interpolation. Our method supports native fisheye inputs and thus laying the groundwork for hybrid reconstruction with different camera types.

\noindent\textbf{2. Cross-View Joint Optimization (CVO):} Our key insight is that overlapping views from different cameras naturally form a stereo setup, providing stronger 3D consistency constraints. Although works like DBARF also establish camera relationships, they mainly relies on feature overlap. For training explicit Gaussian primitives shared across views, we argue that beyond feature overlap, angular diversity is also crucial: Exploiting cross-view correlations can significantly reduce ambiguity by optimizing not only the shape of individual Gaussian but also the distribution of all Gaussians in 3D space. To our knowledge, this idea is absent in concurrent 3DGS works.
This implementation does not rely on COLMAP, and if better methods (e.g.,VGGT) can provide more robust and accurate feature matching information between cameras, these could also be used to further improve camera graph accuracy and thus improve the overall performance. 

In summary, our work (i) introduces the KB projection model for native fisheye input without sacrificing the rasterization’s advantages in 3DGS, and (ii) proposes CVO for improving the overall reconstruction quality especially around view-overlapping areas. Our method is orthogonal to pinhole-based approaches and can be combined with them to further improve performance.

\section{Details of Forward Rendering and Gradient Calculation Algorithm}
\label{sec:algorithm}
\begin{algorithm}
    \caption{\textbf{Forward Rendering}}
    \label{alg:forward}
    \renewcommand{\algorithmicrequire}{\textbf{Input:}}
    \renewcommand{\algorithmicensure}{\textbf{Output:}}
    \begin{algorithmic}[1]
        \Require Gaussian parameters $\{\mu_i, \boldsymbol{\Sigma}_i, c_i, \alpha_i\}_{i=1}^N$; 
        camera extrinsic $W$; camera intrinsic $K$
        \Ensure Rendered fisheye image $I$
        
        \For{each $i \in [1,N]$}
            \State $\mu_{\text{cam}} \gets W^{-1} \mu_i$
            \State $\theta_d \gets \text{Kannala-Brandt}\!\left(\arctan\!\frac{\|\mu_{\text{cam},x,y}\|}{\mu_{\text{cam},z}}\right)$
            \State $\mu_{p_i} \gets \mathcal{F}(\mu_{\text{cam}}, \theta_d)$
            \Comment{$\mathcal{F}$ is the fisheye projection model}
            \State $\boldsymbol{\Sigma}_i^{2D} \gets (\boldsymbol{WJ_\theta})\, \boldsymbol{\Sigma}_i\, (\boldsymbol{WJ_\theta})^\top$
        \EndFor
        \State $I \gets \text{AlphaBlending}(\mu_p, \boldsymbol{\Sigma}^{2D}, c, \alpha)$
        \State \Return $I$
    \end{algorithmic}
\end{algorithm}

\begin{algorithm}
    \caption{\textbf{Backward Gradients Calculation}}
    \label{alg:backward}
    \renewcommand{\algorithmicrequire}{\textbf{Input:}\hfill}
    \renewcommand{\algorithmicensure}{\textbf{Output:}\hfill}
    
    \begin{algorithmic}[1]
        \Require $Render loss:L $, $camera\ extrinsic\ W$, 
        $camera \ intrinsic\ K$
        \Ensure Gradients of Gaussian parameters   
        
        \For{each $i \in [1,N]$}
            \State $\nabla c_i, \nabla \boldsymbol{\Sigma}_i^{2D}, \nabla \mu_{p_i} = \text{Diff}(L)$
            \State $\nabla \boldsymbol{\Sigma}_i = \boldsymbol{(WJ_\theta)}^\top (\nabla \boldsymbol{\Sigma}_i^{2D}) \boldsymbol{(WJ_\theta)}$
            
            \Comment{Gradient propagation to 3D}
             
            \State $\nabla \mu_{i} \leftarrow \text{GradientPropagation}(\nabla \mu_{p_i}, W, K; \nabla c_i; \nabla J_\theta)$
            
            \Comment{Gradient propagation to 3D \& Chain rule}
            
            \State $\nabla s_i, \nabla r_i \leftarrow \text{GradientPropagation}(\nabla \boldsymbol{\Sigma}_i) $

            \Comment{Chain rule to find derivatives w.r.t. scaling and rotation}
            
        \EndFor
        
        \Statex \Return $\nabla \mu, \nabla c, \nabla s, \nabla r$
    \end{algorithmic}
\end{algorithm}

Here, we provide an intuitive mathematical analysis explaining why regions with strong fisheye distortion tend to produce extremely anisotropic Gaussians during optimization. We analyze Gaussian shape optimization by examining the gradients of the covariance matrix as an example.

As shown in Eq.~(7), the fisheye projection Jacobian $\mathbf{J}_\theta$ is a highly nonlinear function of the camera-space coordinates $(x_c, y_c, z_c)$, involving the distorted incident angle $\theta_d$, its derivative $\theta_d'$, and higher-order terms of the radial distance $d$ and depth $z_c$. These factors jointly induce view-dependent scaling and anisotropy in the Jacobian, whose structure varies significantly with the incident angle.

During back-propagation (Algorithm~\ref{alg:backward}), the 3D covariance gradient is given by
$\nabla \boldsymbol{\Sigma}_i
=
(\mathbf{WJ}_\theta)^\top
(\nabla \boldsymbol{\Sigma}_i^{2D})
(\mathbf{WJ}_\theta),$
indicating that the nonlinear Jacobian introduced by the fisheye model directly modulates the covariance update. In regions with large distortion (i.e., large incident angles), the Jacobian exhibits strong view-dependent anisotropy, leading to unbalanced gradients across views. In contrast, near the image center the Jacobian smoothly degenerates toward the pinhole case, resulting in more stable and nearly isotropic optimization.

\begin{figure}
    \centering
    \includegraphics[width=\linewidth]{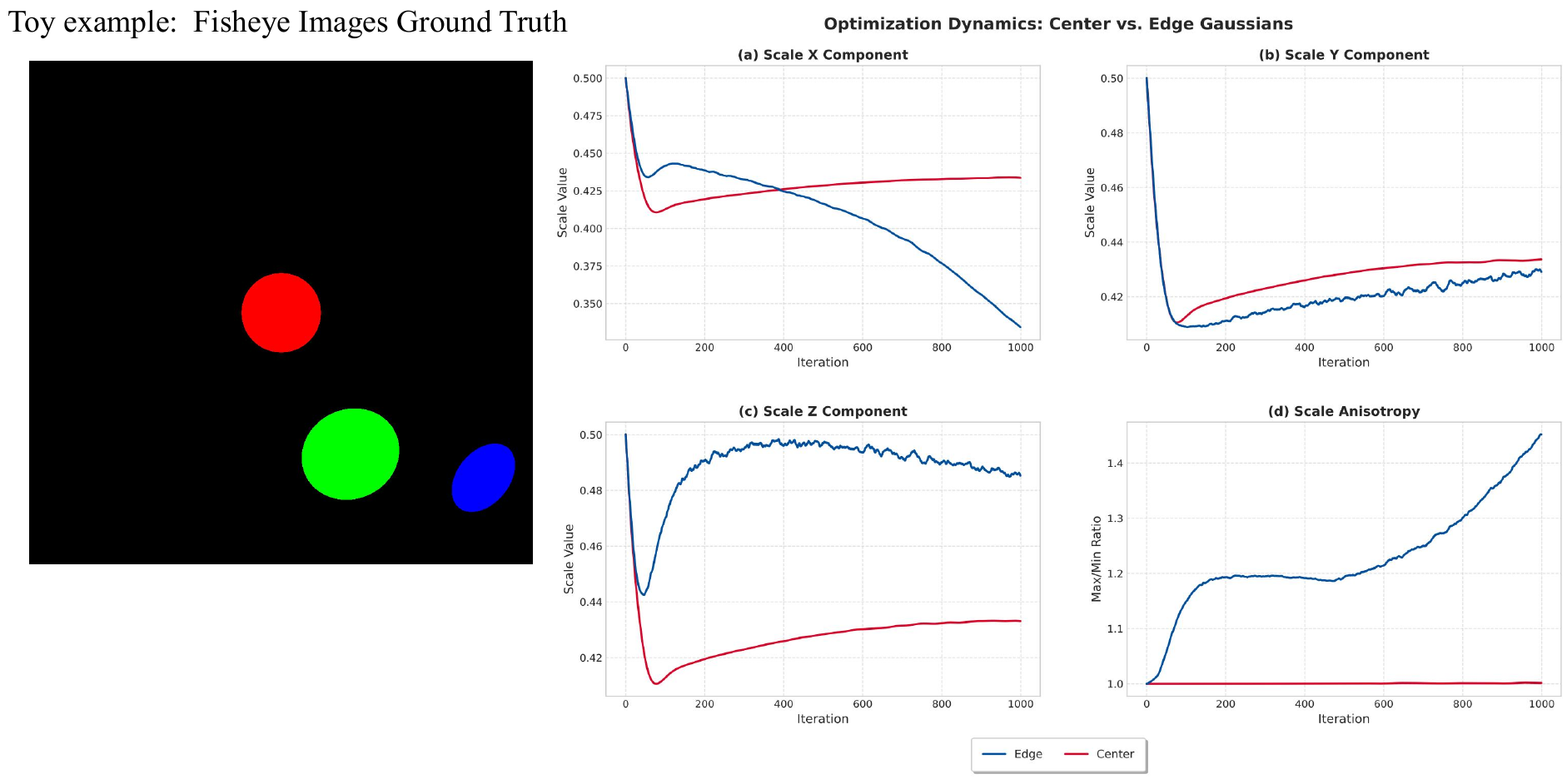}
    \caption{Toy experiment showing that strong fisheye distortion causes unstable and anisotropic Gaussian optimization near image boundaries.}
    \label{fig:math}
\end{figure}

We further illustrate this effect with a toy experiment (Fig.~\ref{fig:math}). Three spheres with identical radii are rendered into multiple fisheye images and fitted using three Gaussian primitives. The results show that the central Gaussian preserves isotropic scale during optimization, whereas the peripheral Gaussian (blue) is pulled by nonlinear, view-dependent gradients, leading to unstable optimization and extreme shape deformation.

These analysis helps explain the optimization instability observed near fisheye image boundaries.

\section{More Quantitative and Qualitative Comparison Results \& Analysis}
\label{sec:more results}
\begin{figure}
    \centering
    \includegraphics[width=\linewidth]{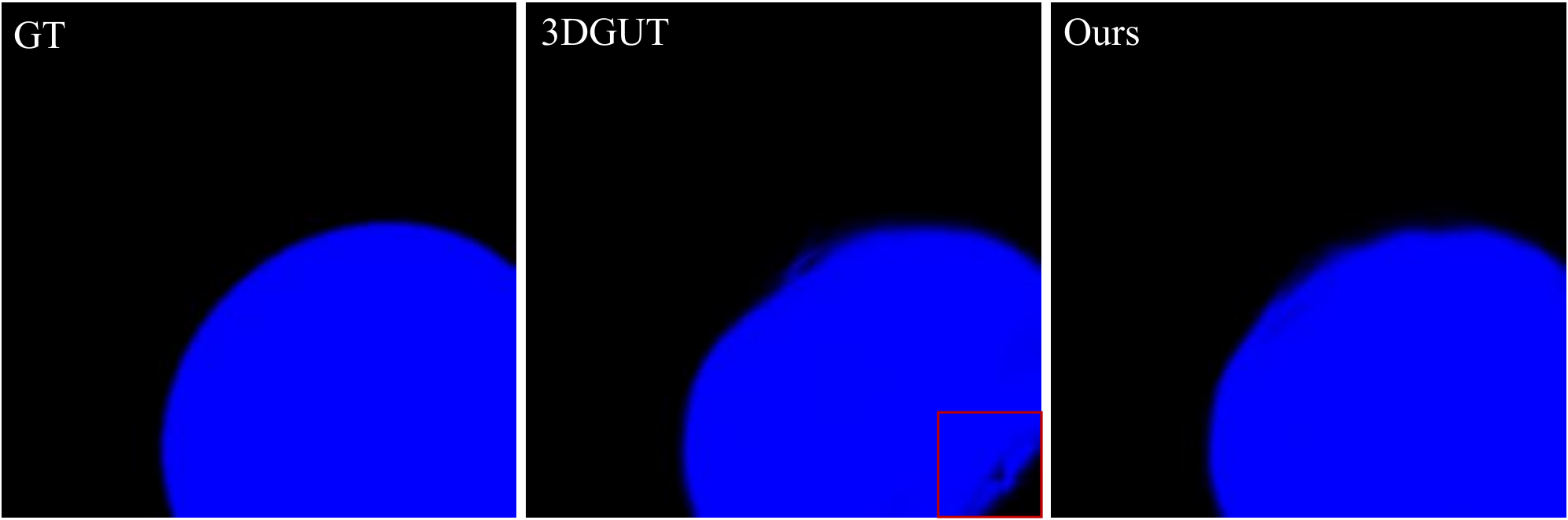}
    \caption{Toy experiment illustrating the behavior of our method and 3DGUT at fisheye image boundaries. We place a blue sphere in 3D space and render multi-view fisheye images as ground truth. Reconstruction results using 3DGUT and our method are shown above. Our method achieves more accurate fitting near the image boundaries and avoids mosaic-like artifacts, while 3DGUT exhibits degraded geometry under strong distortion. }
    \label{fig:3dgut_toy}
\end{figure}

\begin{table*}[h]
\centering
\caption{Per-scene results of our method and related state-of-art works on the FisheyeNeRF~\cite{jeong2021self} Dataset (Test View). \textbf{** Fisheye-GS still need to pre-process data before training.}}
\label{tab:fisheyenerf_test}
\resizebox{\textwidth}{!}{%
\begin{tabular}{lccccccccccccccccccccc}
\toprule
\multicolumn{1}{c}{}                              & \multicolumn{3}{c}{Globe}                                                                         & \multicolumn{3}{c}{Rock}                                                                          & \multicolumn{3}{c}{Cube}                                                                          & \multicolumn{3}{c}{Flowers}                                                                       & \multicolumn{3}{c}{Heart}                                                                         & \multicolumn{3}{c}{Chairs}                                                                        & \multicolumn{3}{c}{\textbf{Average}}                                                                                         \\
\multicolumn{1}{c}{\multirow{-2}{*}{FisheyeNeRF}} & SSIM↑                          & PSNR↑                           & LPIPS↓                         & SSIM↑                          & PSNR↑                           & LPIPS↓                         & SSIM↑                          & PSNR↑                           & LPIPS↓                         & SSIM↑                          & PSNR↑                           & LPIPS↓                         & SSIM↑                          & PSNR↑                           & LPIPS↓                         & SSIM↑                          & PSNR↑                           & LPIPS↓                         & \textbf{SSIM↑}                          & \textbf{PSNR↑}                           & \textbf{LPIPS↓}                         \\ \midrule
\textbf{Fisheye-GS**}                             & \cellcolor[HTML]{FFCB2F}0.7942 & 23.4159                         & \cellcolor[HTML]{FFFE65}0.2667 & \cellcolor[HTML]{FFCB2F}0.7539 & \cellcolor[HTML]{FFFE65}24.4696 & \cellcolor[HTML]{FFFE65}0.2575 & \cellcolor[HTML]{FFCB2F}0.7826 & \cellcolor[HTML]{FFFE65}24.4171 & \cellcolor[HTML]{FFFE65}0.2948 & \cellcolor[HTML]{FFCB2F}0.6843 & \cellcolor[HTML]{FFFE65}21.5505 & \cellcolor[HTML]{FFFE65}0.2901 & \cellcolor[HTML]{FFCB2F}0.7843 & 23.6402                         & \cellcolor[HTML]{FFFE65}0.3108 & 0.6540                         & 18.6131                         & 0.4262                         & \textbf{0.7422}                         & \textbf{22.6844}                         & \cellcolor[HTML]{FFFE65}\textbf{0.3077} \\
\textbf{3DGUT}                                    & 0.7790                         & \cellcolor[HTML]{FFFE65}24.0800 & 0.3420                         & 0.7320                         & 24.1400                         & 0.3130                         & 0.7630                         & 24.3140                         & 0.3690                         & 0.6550                         & 21.3070                         & 0.3690                         & 0.7730                         & \cellcolor[HTML]{FFCB2F}23.6490 & 0.3930                         & \cellcolor[HTML]{FFFE65}0.8520 & \cellcolor[HTML]{FFFE65}24.6950 & \cellcolor[HTML]{FFFE65}0.3070 & \cellcolor[HTML]{FFFE65}\textbf{0.7590} & \cellcolor[HTML]{FFFE65}\textbf{23.6975} & \textbf{0.3488}                         \\
\textbf{Self-Cali-GS}                             & 0.7443                         & 22.1169                         & 0.4273                         & 0.6799                         & 22.2834                         & 0.4777                         & 0.7414                         & 22.4269                         & 0.4573                         & 0.5868                         & 19.5231                         & 0.5212                         & 0.7907                         & 22.3154                         & 0.3788                         & 0.7710                         & 20.9594                         & 0.3535                         & \textbf{0.7190}                               & \textbf{21.6042}                                & \textbf{0.4360}                               \\
\textbf{Ours}                                     & \cellcolor[HTML]{FFFE65}0.7937 & \cellcolor[HTML]{FFCB2F}24.0818 & \cellcolor[HTML]{FFCB2F}0.2423 & \cellcolor[HTML]{FFFE65}0.7480 & \cellcolor[HTML]{FFCB2F}24.5222 & \cellcolor[HTML]{FFCB2F}0.2495 & \cellcolor[HTML]{FFFE65}0.7755 & \cellcolor[HTML]{FFCB2F}24.4438 & \cellcolor[HTML]{FFCB2F}0.2758 & \cellcolor[HTML]{FFFE65}0.6826 & \cellcolor[HTML]{FFCB2F}21.7037 & \cellcolor[HTML]{FFCB2F}0.2776 & \cellcolor[HTML]{FFFE65}0.7798 & \cellcolor[HTML]{FFFE65}23.5701 & \cellcolor[HTML]{FFCB2F}0.2899 & \cellcolor[HTML]{FFCB2F}0.8620 & \cellcolor[HTML]{FFCB2F}25.0001 & \cellcolor[HTML]{FFCB2F}0.2185 & \cellcolor[HTML]{FFCB2F}\textbf{0.7736} & \cellcolor[HTML]{FFCB2F}\textbf{23.8870} & \cellcolor[HTML]{FFCB2F}\textbf{0.2589} \\ \bottomrule
\end{tabular}%
}
\end{table*}

\begin{table*}[]
\caption{Cross-view joint optimization (CVO) under pinhole inputs of Tanks \& Temples~\cite{Knapitsch2017} Dataset.}
\label{tab:CVO-tnt}
\resizebox{\textwidth}{!}{%
\begin{tabular}{lcccccccccccc}
\toprule
                           & \multicolumn{3}{c}{\textbf{drjohnson}}                                 & \multicolumn{3}{c}{\textbf{playroom}}       & \multicolumn{3}{c}{\textbf{train}}                                 & \multicolumn{3}{c}{\textbf{truck}}                                                                                                            \\
\textbf{Method}            & \textbf{SSIM↑} & \textbf{PSNR↑} & \textbf{LPIPS↓} & \multicolumn{1}{c}{\textbf{SSIM↑}} & \multicolumn{1}{c}{\textbf{PSNR↑}} & \multicolumn{1}{c}{\textbf{LPIPS↓}} & \textbf{SSIM↑} & \textbf{PSNR↑} & \textbf{LPIPS↓} & \multicolumn{1}{c}{\textbf{SSIM↑}} & \multicolumn{1}{c}{\textbf{PSNR↑}} & \multicolumn{1}{c}{\textbf{LPIPS↓}}  \\ \midrule
3DGS & 0.9012        & 29.4539       & 0.2448       & 0.9027       & 30.0726       & 0.2482              & 0.8093            & 22.3197                                   & 0.2171       & 0.8780      & 25.6492       & 0.1576                                 \\
3DGS + random select  & 0.8975        & 29.4017       & 0.2434     & 0.9009      & 30.0162       & \textbf{0.2453}                 & 0.8182                                  & 22.3985                                   & 0.2031             & 0.8824      & 25.8986       & 0.1484                           \\
3DGS + CVO         & \textbf{0.9024}      & \textbf{29.5530}       & \textbf{0.2413}        & \textbf{0.9083}      & \textbf{30.3675}       & 0.2466             & \textbf{0.8250}                                  & \textbf{22.7337}                                  & \textbf{0.2010}      &  \textbf{0.8872}     & \textbf{25.9716}       & \textbf{0.1466}                                \\ \bottomrule
\end{tabular}%
}
\end{table*}

\begin{figure*}
    \centering
    \includegraphics[width=0.98\textwidth]{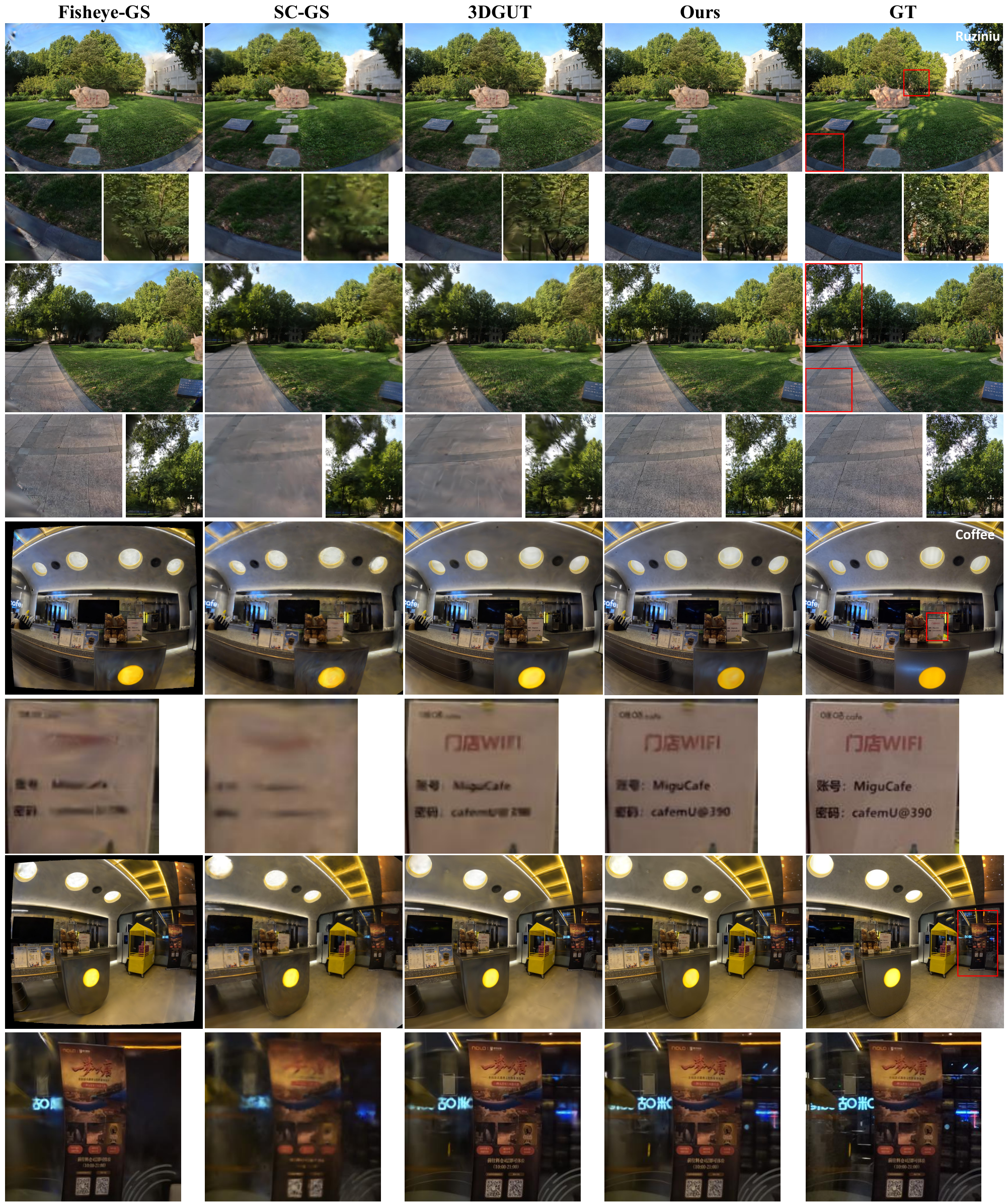}
    \caption{Qualitative comparison on Den-SOFT~\cite{yu2024soft} dataset. Our method achieves the best results in both indoor and outdoor large-scale scenes. }
    \label{fig:den-soft}
\end{figure*}

We present test-view comparison results on the FisheyeNeRF dataset in Table~\ref{tab:fisheyenerf_test} and additional qualitative comparisons on the Den-SOFT dataset in Fig.~\ref{fig:den-soft}, together with further analysis of the performance differences among Fisheye-GS, 3DGUT, SC-GS, and our method.

Fisheye-GS requires preprocessing to convert raw fisheye images into ideal equidistant fisheye projections as ground truth. For datasets with more severe distortions, this preprocessing leads to greater information loss (e.g., visible black borders and reduced effective resolution, stretching-induced loss of high-frequency details). As a result, although Fisheye-GS may obtain slightly higher numerical scores in some cases (e.g., Table~\ref{tab:fisheyenerf}), qualitative results (Fig.~\ref{fig:fisheyenerf} and the supplementary video) better reflect the reconstruction quality, showing noticeable degradation in high-frequency details.

Among the remaining methods, 3DGUT is the closest to ours in performance. For a fair comparison, we follow the same six ScanNet++ scenes used in 3DGUT (Table~\ref{tab:scannet++}). On two compact and highly reflective indoor scenes (IDs: 0a5c013435 and d415cc449b), 3DGUT slightly outperforms our method, which is consistent with its ray-based formulation that is more robust to localized lighting and specular effects and is not the primary focus of this work. In the other scenes and datasets, however, our method consistently achieves higher reconstruction quality.

Although both DirectFisheye-GS and 3DGUT adopt similar nonlinear projection models, their approaches target fundamentally different goals. 3DGUT augments 3DGS with ray tracing via the Unscented Transform (UT), estimating projected Gaussian’s mean and covariance using seven sigma points. This sampling-based formulation improves robustness in light-intensive or reflective scenes, but under strong fisheye distortion, particularly near image boundaries or on datasets with larger distortion such as Den-SOFT, this limited sampling may fail to capture local anisotropy. As a results, the estimated covariances may vary discontinuously across neighboring pixels, leading to blurred details, and mosaic-like artifacts (see Fig.~\ref{fig:fisheyenerf}, Fig.~\ref{fig:den-soft}, and Fig.~\ref{fig:3dgut_toy}, supp.video 2:43, 2:58...).

In contrast, our analytic Jacobian formulation propagates Gaussian covariance through the Kannala–Brandt fisheye model in a continuous manner, enabling more stable covariance updates under large distortion. Combined with the CVO strategy, our method achieves sharper edges, smoother $\alpha$-blending, and better cross-view geometric consistency, showing generality and effectiveness even in dataset like Den-SOFT (Table~\ref{tab:Den-SOFT}, Fig.~\ref{fig:den-soft}), which exhibits more severe distortion, larger spatial coverage, and more challenging in-the-wild conditions (complex lighting, occlusions).

Finally, while numerical metrics between the two methods can be close in some scenes, qualitative results in our figures and videos show that our method preserves finer structures and sharper boundaries, further demonstrating the effectiveness of DirectFisheye-GS.

\section{About Larger Batch-Size}
\label{sec:larger batch-size}
In our experiments, the top-batchsize (defined in Line.332) is set to N = 2. Increasing N improves performance, but it also significantly raises the training time. For example, when N = the number of training images, the view selection strategy becomes irrelevant. We computed the trade-off between performance gain and training time for top-batchsizes N = 1→5 over 30k iterations as shown in Table~\ref{tab:batch_comparison}.

\begin{table}[]
\caption{Performance comparison with varying batch sizes on Scene:Ruziniu in Den-SOFT dataset.}
\label{tab:batch_comparison}
\resizebox{\columnwidth}{!}{%
\begin{tabular}{ccccc}
\toprule
\textbf{Batchsize} & \textbf{PSNR} & \textbf{LPIPS} & \textbf{Time (min)} & \textbf{$\Delta$PSNR/$\Delta$Time} \\
\midrule
1 & 23.7717 & 0.2232 & 38.87 & - \\
2 & 24.0154 & 0.2139 & 62.43 & 1.0344\% \\
3 & 24.1583 & 0.2102 & 89.83 & 0.5215\% \\
4 & 24.2262 & 0.2077 & 129.52 & 0.1711\% \\
5 & 24.2896 & 0.2051 & 160.92 & 0.2019\% \\ \bottomrule
\end{tabular} %
}
\end{table}
While increasing N results in better performance, the most significant improvement happens when moving from N=1→2—shifting from monocular to stereo supervision. This transition is conceptually important.

\section{Computational Cost}
\label{sec:computational cost}
We compare our method with two approaches that support native fisheye input (others require preprocessing). Using the ScanNet++\_8d563fc2cc scene as an example, our method achieves both the best reconstruction quality (refer to Table~\ref{tab:scannet++}) and computational efficiency as shown in Table~\ref{tab:computational-cost}. 
\begin{table}[]
\caption{Comparisons of computational costs for methods supporting native fisheye inputs. All methods are evaluated on a single NVIDIA A100 GPU.}
\label{tab:computational-cost}
\resizebox{\columnwidth}{!}{%
\begin{tabular}{lccc}
\toprule
\textbf{Method} & \textbf{Ours} & \textbf{3DGUT} & \textbf{Self-Cali-GS} \\
\midrule
\text{Training-Time (h)} &\textbf{0.395} & 0.404 & 3.333 \\
\text{Rendering-Speed (FPS)} &\textbf{77} & 55 & 31 \\ \bottomrule
\end{tabular} %
}
\end{table}
\section{Limitations Discussion}
\label{sec:limitations}
A primary limitation is that CVO’s performance gain is not significant under complex lighting conditions. While it benefits optimization of view-dependent attributes (SH) of the same Gaussian, further modeling of complex effects like reflection and refraction is needed under the \textbf{Splat-paradigm}. More precisely model complex reflectance and anisotropy, advanced view-dependent modeling(e.g.,light reflection direction modeling) would be valuable future work.

Another promising work is that CVO uses multi‑view feature correspondences (from COLMAP) to group overlapping views. It improves reconstruction in most areas and does not degrade performance in sparse‑texture regions. However, truly super‑challenging zones are already difficult for 3DGS, calling for improved representation and semantic constraints, which we leave for future work. 

\begin{figure*}
    \centering
    \includegraphics[width=0.92\textwidth]{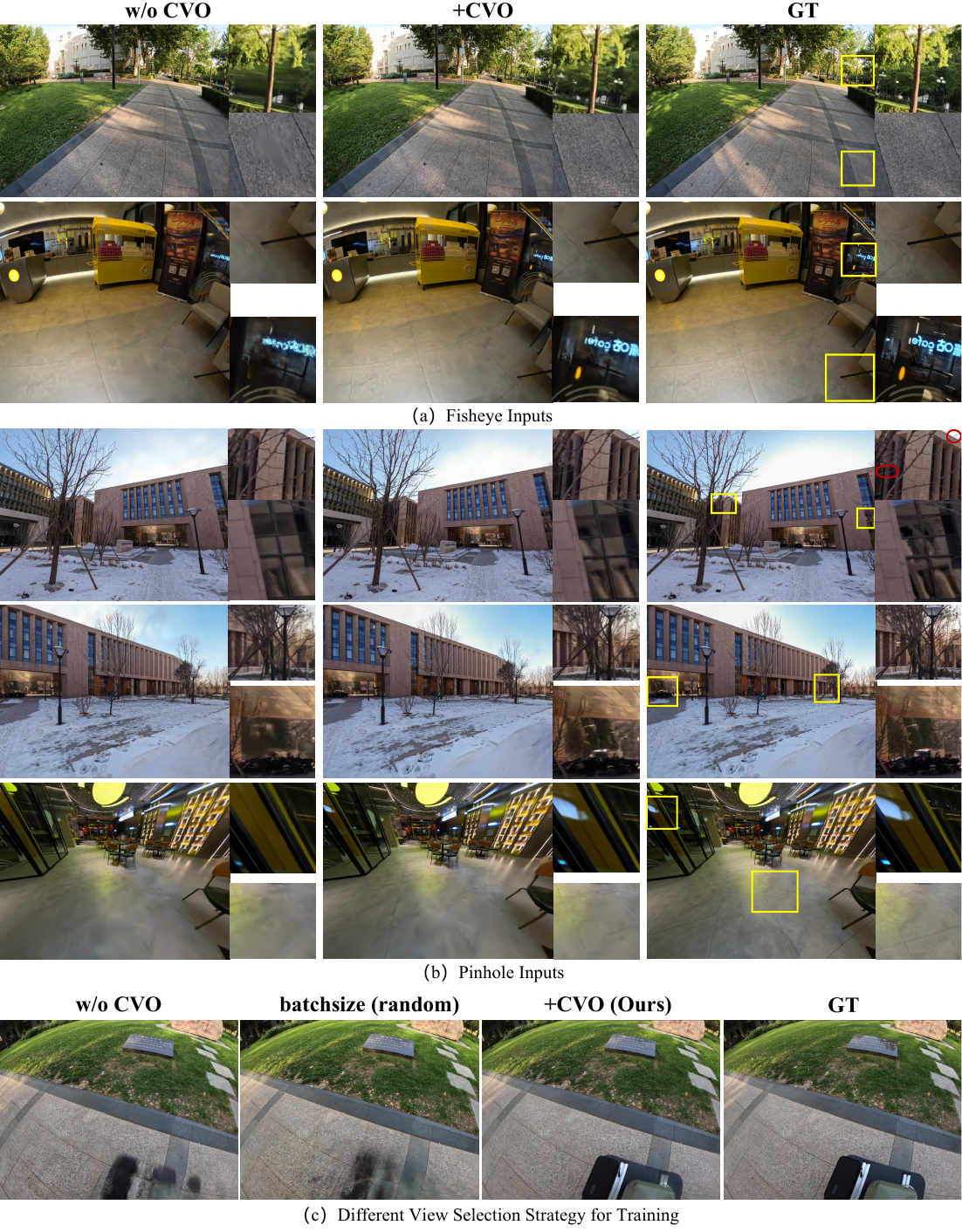}
    \caption{Qualitative comparison of ablation studies on cross-view joint optimization (CVO) with fisheye or pinhole camera inputs.}
    \label{fig:ablation-study}
\end{figure*}

\end{document}